%% file: acl2023.tex
\pdfoutput=1

\PassOptionsToPackage{svgnames,dvipsnames}{xcolor}

\documentclass[11pt]{article}

\usepackage[]{ACL2023}

\usepackage{times}
\usepackage{latexsym}

\usepackage{algorithm}
\usepackage{algorithmic}
\usepackage{amsmath}

\usepackage{makecell}
\usepackage[most]{tcolorbox}
\usepackage{booktabs}
\usepackage{graphicx}
\usepackage{caption}
\usepackage{adjustbox}
\usepackage{multirow}
\usepackage{colortbl}
\usepackage{lscape}
\usepackage{array}
\usepackage{courier}
\usepackage{enumitem}
\usepackage{pifont}
\usepackage{placeins}

\usepackage[T1]{fontenc}

\usepackage[utf8]{inputenc}

\usepackage{microtype}

\usepackage{inconsolata}

\usepackage{booktabs}

\usepackage{graphicx}

\usepackage{ragged2e}

\newcommand\datasetname{\textcolor{black}{\textsc{Unggah-Ungguh}}}

\newcommand{\honorificBox}[2]{%
  \tcbox[colback=#1!20,
         colframe=#1,
         arc=5pt,
         boxrule=0.8pt,
         left=2pt,
         right=2pt,
         top=1pt,
         bottom=1pt,
         on line]{\large{\textit{#2}}}%
}

\newcommand{\levelIndicator}[1]{%
  \raisebox{0.5ex}{%
    \tcbox[colback=#1,
           colframe=#1,
           arc=0pt,
           boxrule=0pt,
           left=2pt,
           right=2pt,
           top=0pt,
           bottom=0pt,
           width=3mm,
           height=2ex,
           valign=center,
           on line]{}%
  }%
  \space%
}

\newcommand{\cmark}{\tcbox[colback=Green!30,colframe=Green,arc=4pt,boxrule=0pt,left=0.5pt,right=0.5pt,top=0.5pt,bottom=0.5pt, on line]{\ding{51}}} 
\newcommand{\xmark}{\tcbox[colback=Gray!20,colframe=Gray,arc=4pt,boxrule=0pt,left=0.5pt,right=0.5pt,top=0.5pt,bottom=0.5pt, on line]{\ding{55}}} 

\usepackage{amsmath}

\definecolor{bestcolor}{HTML}{DEACF5}
\definecolor{lightgray}{gray}{0.85}

\title{Do Language Models Understand Honorific Systems in Javanese?}

\author{
    \textbf{Mohammad Rifqi Farhansyah}$^{\alpha\beta}$\thanks{$\text{ }$ Equal contribution.} \quad
    \textbf{Iwan Darmawan}$^{\beta}$\footnotemark[1] \quad
    \textbf{Adryan Kusumawardhana}$^{\beta}$ \quad \\
    \textbf{Genta Indra Winata}$^{\gamma\dagger}$\thanks{$\text{ }$ The work was conducted outside Capital One.}\thanks{$\text{ }$ Senior authors.} \quad
    \textbf{Alham Fikri Aji}$^{\delta}$\footnotemark[3] \quad
    \textbf{Derry Tanti Wijaya}$^{\beta\zeta}$\footnotemark[3] \\
    $^\alpha$Institut Teknologi Bandung \quad
    $^\beta$Monash Indonesia \quad
    $^\zeta$Boston University \quad
    $^\gamma$Capital One \quad
    $^\delta$MBZUAI \\
    \texttt{\{mrifqifarhansyah, adryan.kusumawardhana\}@gmail.com} \quad
    \texttt{idar0006@student.monash.edu} \\
    \texttt{genta.winata@capitalone.com} \quad
    \texttt{alham.fikri@mbzuai.ac.ae} \\
    \texttt{derry.wijaya@monash.edu} \\
}

\begin{document}
\maketitle
\begin{abstract}

The Javanese language features a complex system of honorifics that vary according to the social status of the speaker, listener, and referent. Despite its cultural and linguistic significance, there has been limited progress in developing a comprehensive corpus to  capture these variations for natural language processing (NLP) tasks. In this paper, we present $\datasetname$\footnote{We released our models \& dataset in \url{https://huggingface.co/JavaneseHonorifics}}, a carefully curated dataset designed to encapsulate the nuances of
\textit{Unggah-Ungguh Basa}, the Javanese speech etiquette framework that dictates 
the choice of words and phrases based on social hierarchy and context. Using $\datasetname$, we assess the ability of language models (LMs) to process various levels of Javanese honorifics through classification and machine translation tasks. To further evaluate cross-lingual LMs, we conduct machine translation experiments between Javanese (at specific honorific levels) and Indonesian. Additionally, we explore whether LMs can generate contextually appropriate Javanese honorifics in conversation tasks, 
where the honorific usage should 
align with the social role and contextual cues. Our findings indicate that current LMs struggle with most honorific levels, exhibiting
a bias toward certain honorific tiers.
\end{abstract}

\section{Introduction}
In many languages, honorifics play a crucial role in  addressing others and maintaining 
social relationships~\cite{agha1994honorification}. Honorific registers are formally discrete yet functionally stratified systems, meaning that a seemingly fixed set of linguistic forms enables speakers to navigate multiple aspects of the pragmatic context in which language is used~\cite{agha1998stereotypes}. Using the appropriate honorific registers in the right situations fosters culturally appropriate conversations and enhances the overall propriety of interactions. One language that exemplifies a highly complex honorific system is Javanese, where honorific usage is deeply intertwined with cultural norms and historical traditions. In Javanese, the use of honorific language is inseparable from the broader concept of linguistic politeness~\cite{rahayu2014comparison}.

\begin{table}
\renewcommand{\arraystretch}{1.5}
\centering
\resizebox{0.49\textwidth}{!}{
\begin{tabular}{ll}
\toprule
\levelIndicator{RoyalBlue}\textbf{Ngoko} & Mbak Srini \honorificBox{RoyalBlue}{mangan} pecel \honorificBox{RoyalBlue}{ajange} pincuk. \\
\levelIndicator{Purple}\textbf{Ngoko Alus} & Mbak Srini \honorificBox{Purple}{dhahar} pecel \honorificBox{Purple}{ambenge} pincuk.\\
\levelIndicator{SeaGreen}\textbf{Krama} & Mbak Srini \honorificBox{SeaGreen}{nedha} pecel \honorificBox{SeaGreen}{ajangipun} pincuk. \\
\levelIndicator{Salmon}\textbf{Krama Alus} & Mbak Srini \honorificBox{Salmon}{dhahar} pecel \honorificBox{Salmon}{ambengipun} pincuk.\\ 
\midrule
\rowcolor{lightgray}\textbf{Translation \raisebox{-0.25em}{\includegraphics[height=1em]{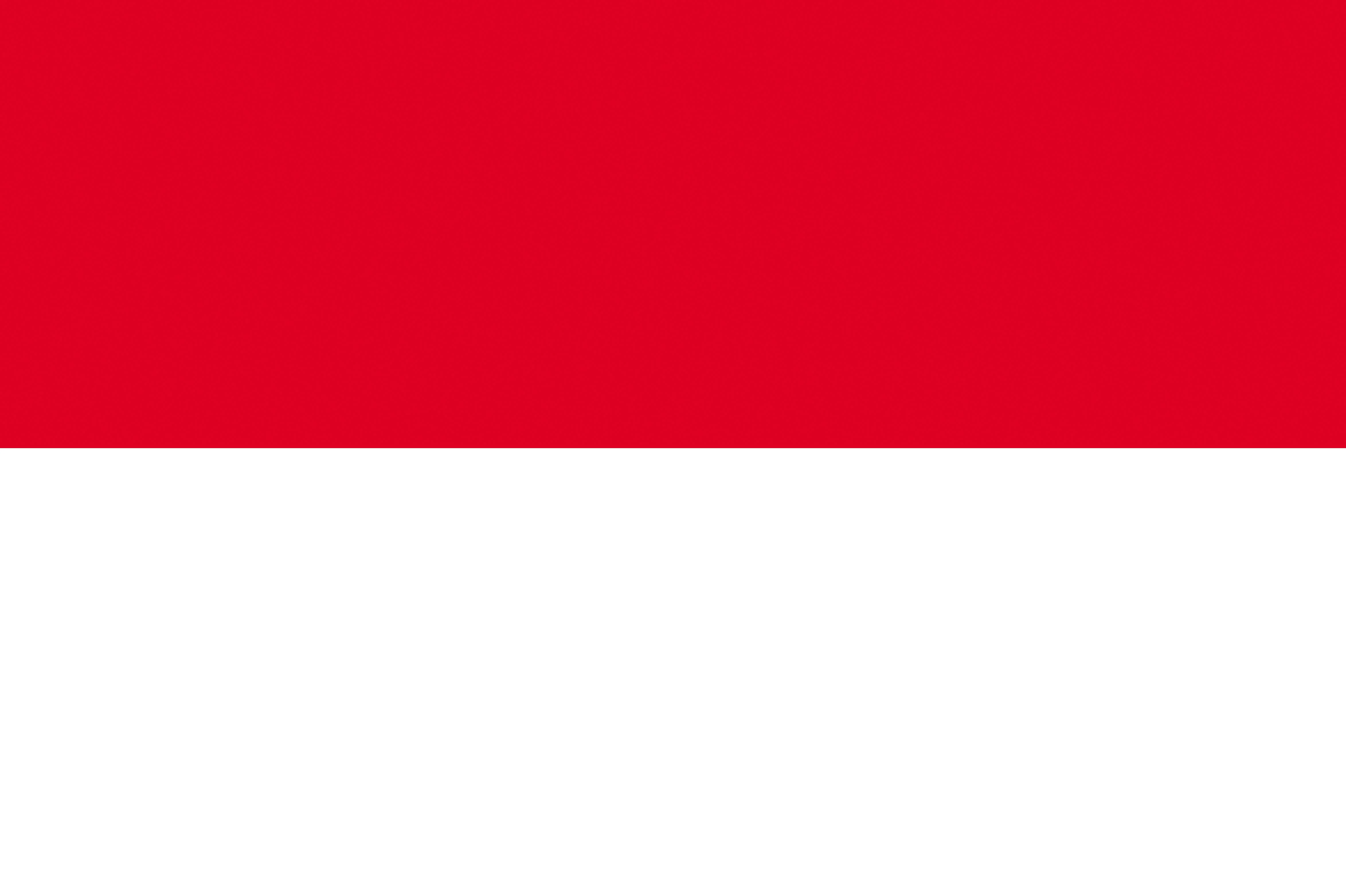}}} & Mbak Srini \textit{makan} pecel \textit{dengan ajang} pincuk. \\
\midrule 
\rowcolor{lightgray}\textbf{Translation \raisebox{-0.25em}{\includegraphics[height=1em]{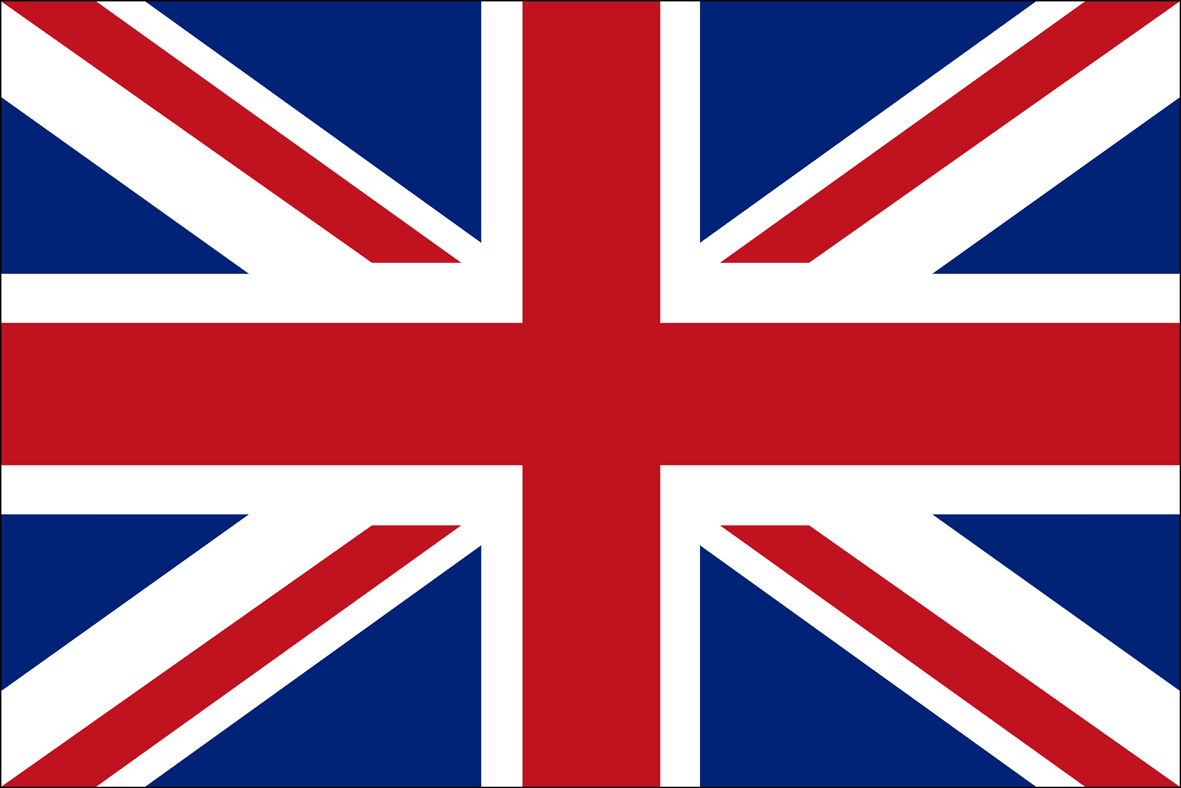}}} & Mbak Srini \textit{ate} pecel \textit{with} pincuk. \\
\midrule 
\levelIndicator{RoyalBlue}\textbf{Ngoko} & Bu Haiti \honorificBox{RoyalBlue}{arep tuku} gula selawe kilo, \honorificBox{RoyalBlue}{wis dak-dolane}.\\
\levelIndicator{Purple}\textbf{Ngoko Alus} & Bu Haiti \honorificBox{Purple}{arep mundhut} gula selawe kilo, \honorificBox{Purple}{wis adalem dolane}\\
\levelIndicator{SeaGreen}\textbf{Krama} & Bu Haiti \honorificBox{SeaGreen}{badhe tumbas} gendhis selangkung kilo, \honorificBox{SeaGreen}{sampun}\\
& \honorificBox{SeaGreen}{kula sadeanipun}. \\
\levelIndicator{Salmon}\textbf{Krama Alus} & Bu Haiti \honorificBox{Salmon}{badhe mundhut} gendhis selangkung kilo, \honorificBox{Salmon}{sampun} \\
& \honorificBox{Salmon}{adalem sadeanipun}. \\
\midrule
\rowcolor{lightgray}\textbf{Translation \raisebox{-0.25em}{\includegraphics[height=1em]{img/Indonesia-flag.png}}} & Bu Haiti \textit{akan membeli} gula dua puluh lima kilo, \textit{sudah saya jualnya}. \\ 
\midrule
\rowcolor{lightgray}\textbf{Translation \raisebox{-0.25em}{\includegraphics[height=1em]{img/UK-flag.png}}} & Mrs. Haiti \textit{is going to buy} twenty-five kilos of sugar, \textit{I have already sold it}. \\ 
\bottomrule
\end{tabular}
}
\caption{$\datasetname$ examples with four different honorific levels from the most informal (Ngoko) to the most refined (Krama Alus). 
The \textit{italicized text} highlights the phrases that vary according to the honorific level. More examples can be found in Appendix \ref{app:unggahungguhsamples}.}
\label{examplesUU}
\renewcommand{\arraystretch}{1.0}
\vspace{-1em}
\end{table}

\begin{figure*}[!th]
    \centering
    \small
    \includegraphics[width=\textwidth]{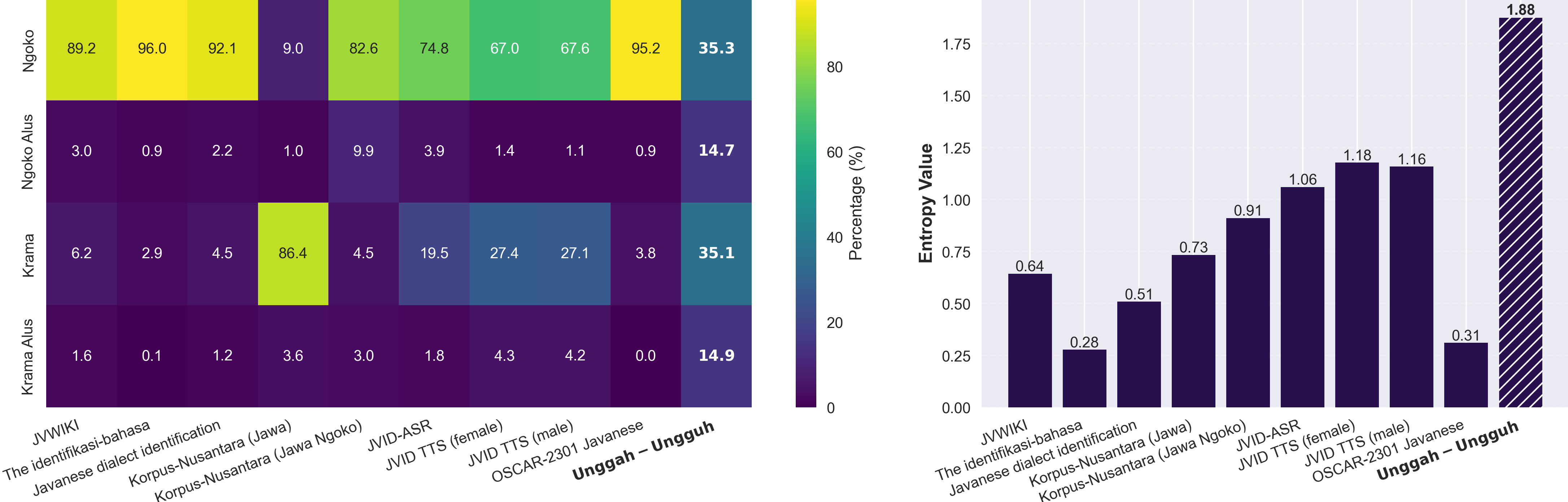}
    \caption{A heatmap \textbf{(left)} illustrates the distribution (\%) of Javanese honorific levels across various corpora including ours (Unggah-Ungguh). An accompanying entropy bar chart \textbf{(right)} quantifies the variability within each corpus using Shannon's Entropy. A higher entropy value indicates a more diverse and well-balanced corpus.}
    \label{fig:dataset_coverage}
    \vspace{-1em}
\end{figure*}

The Javanese language, spoken by over 98 million people,\footnote{\url{https://www.langcen.cam.ac.uk/resources/langj/javanese.html}.} features a distinctive honorific system known as \textit{Unggah-Ungguh Basa}\footnote{More details about Unggah-Ungguh Basa can be seen in Appendix \ref{app:unggahungguhbasa}}, which is essential for conveying respect, social hierarchy, and formality in conversations. The four primary levels of Javanese honorifics (Table \ref{examplesUU})—Ngoko, Ngoko Alus, Krama, and Krama Alus—each represent different degrees of formality and respect, playing a crucial role in facilitating appropriate social interactions within Javanese society. Despite their cultural significance, there is a notable lack of comprehensive linguistic resources that accurately capture these distinctions for natural language processing (NLP) applications. 

Current research indicates that existing models struggle to accurately interpret and generate Javanese honorifics due to the absence of a well-annotated corpus \citep{Marreddy2022AmIA}, which in turn hinders the development of effective NLP tools capable of handling their complexity. Moreover, as illustrated in Figure~\ref{fig:dataset_coverage}, most existing Javanese corpora exhibit an imbalanced distribution of honorific levels, further limiting model performance. Addressing this issue is critical, particularly as language models increasingly serve as personal assistants across various domains, adapting to user expectations~\cite{kong-etal-2024-better}. Given that honorifics shape social interactions, an LM’s ability to process and generate them appropriately is essential for maintaining perceived status and formality, fostering trust and likability~\cite{JEON2022892}, and ultimately enabling more natural and culturally sensitive human-AI interactions.

In this paper, we aim to bridge the existing gap by systematically evaluating the ability of LMs to comprehend and generate appropriate Javanese honorifics. We conduct a series of evaluations to identify potential biases toward specific honorific levels, assess cross-lingual performance in handling Javanese honorifics, and analyze models' ability to generate conversational text with contextually appropriate honorific usage based on the speaker’s role and context. Our contributions are:

\begin{itemize}[noitemsep, topsep=0pt, partopsep=0pt, parsep=0pt]
    \item We introduce $\datasetname$, the first multi-cultural honorific corpus for the Javanese language. Each sentence in this corpus is annotated with one of the four honorific levels.
    \item We leverage this corpus for four downstream NLP tasks: honorific-level classification, honorific style change, cross-lingual honorific translation, and conversation generation with honorific personas.
    \item We explore a diverse set of LMs to address these tasks, including English-centric, multilingual, and Southeast Asian regional models, as well as models trained on Indonesian and its regional languages. Our approaches include encoder-based models, generative models, and rule-based methods.
\end{itemize}
Beyond advancing NLP research, this corpus serves as a vital resource and an important first step toward accurately digitizing the nuanced aspects of Javanese cultural heritage. By making it available, we aim to support the development of more accurate and culturally sensitive NLP models for the Javanese language while also encouraging future research on other low-resource languages with similarly complex sociolinguistic structures.

\section{Related Work}
\begin{table*}[!h]
\centering
\resizebox{\textwidth}{!}{
\begin{tabular}{lllll}
\toprule
\textbf{Source} & \textbf{Title} & \textbf{Author/Editor} & \textbf{Publication Year} & \textbf{Publisher} \\
\midrule
Dictionary & Kamus Unggah-Ungguh Basa Jawa & Drs. Haryana Harjawiyana, S.U. \& Drs. Th. Supriya & 2009 & PT Kanisius \\
\midrule
Book & Unggah-ungguh Basa Jawa & Umi Kuntari, S.S. & 2017 & Pustaka Widyatama \\
 & Mempelajari Unggah-Ungguh Bahasa Jawa & Drs. Imam Riyadi, M.KPd. & 2019 & Penerbit PARAMARTA Trenggalek \\
 & Unggah-ungguh Basa & H. Soemardi & 2016 & CV Satubuku \\
\bottomrule
\end{tabular}
}
\caption{List of all data sources used to construct $\datasetname$.}
\label{tab:data_sources}
\end{table*}

Research on low-resource languages, particularly Javanese, has made significant progress, making it highly relevant to this study. In the domain of language-specific modeling and processing, work on Javanese language modeling~\cite{Wongso2021, cahyawijaya2024cendol, owen2024komodo}, machine translation~\cite{sujaini2020improving, wibawa2013indonesian, wibawa2013hybrid}, part-of-speech tagging~\cite{pratama2020part, ramadhan2020part, noor2020part, enrique2024javanese}, sentiment analysis~\cite{winata2023nusax, lucky2023unveiling}, and dialect identification~\cite{Hidayatullah2020, filby2024dialect} have addressed some of the inherent complexities in the Javanese language. Furthermore, research on Javanese dependency parsing~\cite{Ghiffari2023} has further enhanced our understanding of Javanese linguistic structures. These studies provide a foundation for developing specialized corpora that capture Javanese honorifics.

Dataset creation and curation play a crucial role in advancing research on low-resource languages. Notable efforts in Javanese include \citet{Wibawa2018}, who developed Javanese and Sundanese corpora through community collaboration, leveraging local expertise to build comprehensive datasets. The Japanese Honorific Corpus by \citet{Liu2022} provides a valuable comparative framework for integrating honorific distinctions into linguistic corpora, offering insights applicable to the Javanese context.

Ensuring data quality and rigorous evaluation is essential, particularly for developing reliable and culturally relevant corpora. Previous studies on human evaluation of web-crawled corpora~\cite{Ramírez-Sánchez2022} and dataset audits~\cite{Kreutzer2022} emphasize the importance of stringent quality control measures in corpus development. Collectively, these studies inform our approach to constructing a Javanese honorific corpus, highlighting the need for culturally informed data collection, meticulous annotation, and robust evaluation methods. Drawing from these insights, the development of a comprehensive corpus for the Javanese honorific system based on\textit{ Unggah-Ungguh Basa} can significantly enhance the accuracy and cultural relevance of NLP applications in Javanese.

\begin{table*}[!th]
\centering
\resizebox{\textwidth}{!}{
\begin{tabular}{lrrrrr}
\toprule
\textbf{Text Label} & \textbf{\# Sentences} & \textbf{Avg. Sentence Length} & \textbf{\# Word Tokens} & \textbf{\# Word Types} & \textbf{Yule’s characteristic $K$} 
\\
\midrule
$\quad$Ngoko & 1,419 & 9.26 & 13,142 & 3,486 & 118.80 
\\
$\quad$Ngoko Alus & 590 & 10.07 & 5,944 & 1,527 & 108.13 
\\
$\quad$Krama & 1,414 & 9.60 & 13,572 & 3,280 & 124.61 
\\
$\quad$Krama Alus & 601 & 10.13 & 6,088 & 1,530 & 115.14 
\\ \midrule
\textbf{Overall} & 4,024 & 9.63 & 38,746 & 6,156 & 105.43 
\\ 
\bottomrule
\end{tabular}
}
\caption{Data Statistics of $\datasetname$.}
\label{tab:stat}
\vspace{-1em}
\end{table*}

\section{$\datasetname$ Corpus}

The $\datasetname$ Corpus is a meticulously curated dataset focused on capturing the nuanced use of honorific language in Javanese.\footnote{Our dataset will be licensed under CC-BY-NC 4.0 for research purposes only.} The dataset ensures comprehensive coverage of honorific usage by drawing from a diverse range of reputable sources, including dictionaries and books. This diverse sourcing results in a context-rich collection.

\subsection{Honorific System}
The Javanese language features a distinctive system of speech levels known as \textit{undha usuk}, which reflects the social hierarchy and relationships between speakers. Traditionally, this system was divided into various levels; however, in modern times, it has been streamlined into four primary levels: \textit{Ngoko}, \textit{Ngoko Alus}, \textit{Krama}, and \textit{Krama Alus}. Each level possesses unique linguistic characteristics and specific contexts of use, making them essential for accurate annotation and analysis in the development of a Javanese honorific corpus.
\paragraph{Ngoko.}
Ngoko is the most informal speech level, used in everyday communication among equals or when addressing someone of lower social status. It consists entirely of ngoko words.

\paragraph{Ngoko Alus.}
Ngoko Alus is a refined version of ngoko, incorporating polite or respectful words. It is typically used when speaking to someone of slightly higher status or to show a degree of respect while maintaining the informal structure of ngoko.

\paragraph{Krama.}
Krama is a more formal speech level, used when addressing someone of higher social status or someone unfamiliar. It is characterized by the use of distinct krama vocabulary, which differs significantly from ngoko.

\paragraph{Krama Alus.}
Krama Alus is the most refined and polite speech level, reserved for interactions with individuals of significantly higher social status or in very formal settings. It involves the use of additional honorific words and phrases that further elevate the politeness of speech.

\subsection{Corpus Creation}
Table~\ref{tab:data_sources} summarizes the main sources used to construct the Javanese Honorific Corpus. The dataset primarily originates from Kamus Unggah-Ungguh Basa Jawa~\cite{harjawiyana2001kamus}, a dictionary containing example sentences labeled by honorific level. Since the source is not digitized, we scanned relevant pages, applied OCR, and manually corrected errors through a two-stage verification process involving native Javanese speakers.

In the first stage, one of the authors (native Javanese speaker) corrected OCR errors by comparing outputs with the original text. In the second stage, another native speaker independently reviewed all sentences and transcribed Indonesian translations, identifying and fixing 58 errors (1.5\%) out of 4,024 sentences. Additional example sentences were sourced from other Javanese language references and manually verified to ensure accuracy and relevance in representing Javanese honorifics.

\subsection{Data Statistics}

Table \ref{tab:stat} provides a statistical overview of the Javanese honorific corpus, comprising 4,024 sentences across four honorific levels: Ngoko (1,419 sentences), Ngoko Alus (590), Krama (1,414), and Krama Alus (601). The $\datasetname$ corpus shows significant variability and a relatively balanced distribution compared to other datasets, as indicated by a Shannon entropy\textsuperscript{\ref{eq:shannon_entropy}} value of 1.88, surpassing nine other datasets, as illustrated in Figure~\ref{fig:dataset_coverage}. This result highlights the corpus's balanced diversity across different honorific levels, making it a valuable resource for studying linguistic diversity in Javanese.

The average sentence length varies slightly across honorific levels, ranging from 9.26 to 10.13 words, reflecting the increased linguistic complexity at higher levels. The corpus contains 38,746 word tokens and 6,156 unique word types, with Yule’s characteristic $K$\textsuperscript{\ref{eq:yule-k}} value \citep{YuleG.Udny1944Tsso} calculated as 105.43.

Yule's $K$ measures word repetition, with higher values indicating more frequent repetition (lower lexical diversity) and lower values reflecting less repetition (higher lexical diversity). In this study, our Javanese Honorific Corpus has a Yule's $K$ value of 105.43, compared to 125.54 in the Japanese Honorific Corpus \citep{Liu2022}, indicating that the Javanese Honorific Corpus exhibits greater lexical diversity than its Japanese counterpart.

\subsection{Tasks}

In our benchmark, we want to train and evaluate LMs' understanding to various language styles in different honorific levels. Our benchmark comprises four downstream NLP tasks:
\paragraph{Task 1: Honorific Level Classification.}
The task is to classify a text into one of the honorific levels. Given an input text $x$, we intend to use some LMs $\theta$ to map $x$ to an honorific-level label $\hat{h}$ from four honorific levels. This evaluation is crucial for assessing the capability of LMs in recognizing specific honorific level within Javanese sentence.
\paragraph{Task 2: Honorific Style Change.} The task is to style translate from a given text $x$ from a source honorific style $h_{src}$ to a target honorific style $h_{tgt}$ in Javanese, resulting a translated text $\hat{y}$. This task is particularly important for evaluating model's capability to capture nuanced sociolinguistic conventions and generate contextually appropriate text across different levels of politeness and respect.

\begin{table*}[!th]
\centering
\resizebox{\textwidth}{!}{
    \begin{tabular}{lcccc|c}
    \toprule
    & Ngoko-Indonesia & Ngoko Alus-Indonesia & Krama-Indonesia & Krama Alus-Indonesia & \textbf{Overall} \\
    \midrule
    KL-Divergence & $1.481$ & $1.29$ & $1.72$ & $1.38$ & $2.26$ \\
    Jensen Score & $0.25 \pm 0.00007$ & $0.23 \pm 0.00014$ & $0.28 \pm 0.00007$ & $0.24 \pm 0.00014$ & $0.34 \pm 0.00005$ \\
    \bottomrule
    \end{tabular}
}
\caption{
KL Divergence and Jensen Score between Javanese honorific-level distributions and their Indonesian translations in $\datasetname$.
}
\label{tab:kl-divergence-jensen}
\vspace{-1em}
\end{table*}

\paragraph{Task 3: Cross-lingual Honorific Translation.} The task is to translate a given sentence $x$ from a specific honorific level $h$ to Indonesian (X $\rightarrow$ ID), $\hat{y}$, and vice versa (ID $\rightarrow$ X). This task is crucial for evaluating how well LMs handle cross-lingual translation between languages with asymmetrical honorific systems (e.g., Javanese (rich honorific system) and Indonesian (lacks explicit honorifics). Furthermore, KL Divergence\textsuperscript{\ref{eq:kl-divergence}}\citep{kullback1951information} and Jensen Score\textsuperscript{\ref{eq:jensen-score}}\citep{lin2002divergence} values computed from $\datasetname$ (Table~\ref{tab:kl-divergence-jensen}) between Indonesian and Javanese indicate a substantial overall divergence, with scores reaching 2.26 and 0.34 respectively. Higher divergence values indicate greater lexical shifts during cross-lingual translation, with Jensen Score reflecting mean distributional distance and KL Divergence capturing asymmetrical divergence between token distributions. These values suggest significant lexical shifts between Javanese and Indonesian. Consequently, the cross-lingual honorific translation task becomes particularly challenging, as models must effectively bridge this lexical gap while preserving both semantic content and sociolinguistic nuance.
\paragraph{Task 4: Conversation Generation.} In this task, we aim to simulate and synthesize conversations between two speakers to assess whether LMs can generate appropriate honorific-specific language based on the speakers' social status. Given the social statuses (e.g., student and teacher) of speakers A and B, respectively, along with a context $c$, the objective is to generate a conversation $\hat{y}$ featuring one utterance per speaker. Each utterance should correspond to the appropriate honorific level dictated by $h_A$ and $h_B$, while maintaining coherence with $c$. For instance, in a conversation between a student and a teacher, the student is expected to speak in a more formal register (Krama Alus), whereas the teacher might use a more casual style (Ngoko). We manually curate 160 different evaluation scenarios, detailed in Appendix \ref{app:conversationrole}. Evaluating this task is crucial for measuring cultural appropriateness in scenarios involving Javanese speakers. 

\section{Experimental Setup}
We describe the experimental\footnote{Our code are available at \url{https://github.com/JavaneseHonorifics/Unggah-Ungguh}} details to train and evaluate models for classifying Javanese honorific levels, as well as the experiments conducted for the four honorific-related tasks. Our approach involves using encoder LMs, generative models, and a rule-based model. We utilize two categories of models: \textit{Fine-tuned Models} and \textit{Off-the-Shelf Models}. Fine-tuned models are specifically employed for Task 1, Honorific Level Classification, with the best-performing classifier subsequently used to evaluate the accuracy of the generated text's honorific level in Task 4, Conversation Generation. Details regarding computational resources and model hyperparameters are provided in Appendix \ref{sec:hyper-parameters}.

\subsection{Fine-tuned Models}

We fine-tune encoder-based and decoder-based Javanese LMs to classify Javanese sentences into specific honorific levels. Specifically, we fine-tune two encoder-based LMs on $\datasetname$: Javanese BERT and Javanese DistilBERT, both of which are pre-trained on Javanese text corpora~\cite{wongso2021causal}. Additionally, we fine-tune a decoder-based LM, Javanese GPT-2, originally designed for movie classification using Javanese IMDB reviews.  

To establish baselines, we fine-tune an LSTM model and develop a rule-based classifier. The rule-based model is implemented using a dictionary-based approach, mapping word types to their corresponding honorific levels based on predefined linguistic rules, as outlined in Algorithm \ref{alg:javanese_classification}. The dictionary used in this model is derived from the ``Indonesian Javanese Dictionary Starter Kit''~\citep{rahutomo2018indonesian}, which categorizes Javanese words into three types: \textit{Ngoko}, \textit{Krama Alus}, and \textit{Krama Inggil}. This dataset 
enables the mapping of Javanese word types to their respective honorific levels. For fine-tuning, we use 80\% of our $\datasetname$ for training, while the remaining 20\% is reserved for evaluating the honorific level classification models and selecting the best-performing classifier.  

\subsection{Off-the-Shelf Models}

We evaluate various models across four tasks, encompassing both closed-source and open-source options, and categorize them into: English-centric models, multilingual models, Southeast Asian regional models, and models fine-tuned specifically for Indonesian and local languages (eg. Javanese, Sundanese, etc.).

For closed-source models, we utilize GPT-4o~\cite{achiam2023gpt} and Gemini 1.5 Pro~\cite{team2024gemma}, noted for their outstanding performance and status as state-of-the-art language models. In the open-source category, we employ Gemma 2 9B Instruct~\cite{team2024gemma} as English-centric model and Llama 3.1 8B Instruct~\cite{dubey2024llama} as multilingual model. These models were chosen for their high performance and serve as benchmarks for the SahabatAI~\footnote{\url{https://huggingface.co/GoToCompany/llama3-8b-cpt-sahabatai-v1-instruct}}~\footnote{\url{https://huggingface.co/GoToCompany/gemma2-9b-cpt-sahabatai-v1-instruct}} models, which are designed for Indonesian contexts. Additionally, we include the Southeast Asian regional model, Sailor2 8B~\cite{dou2024sailoropenlanguagemodels}, as a comparative baseline.

\paragraph{Task Setup.} For all experiments, we run once for each task and setting.
We describe the experimental setting for each task as following:

\begin{enumerate} [noitemsep, topsep=0pt, partopsep=0pt, parsep=0pt] 
\item \textbf{Zero-shot classification.} We classify Javanese sentences into specific honorific levels using zero-shot approach. This method is applied to Task (1) and compared against the fine-tuned models.

\item \textbf{Zero-shot translation.} In Tasks (2) and (3), we use off-the-shelf models to perform translation between honorific levels using a zero-shot approach.  

\item \textbf{One-shot generation.} For Task (4), Conversation Generation, we employ a one-shot approach in which off-the-shelf models are provided with a sample conversation as a formatting guide. Furthermore, we conduct experiments with explicit hints about Javanese honorific usage to assess whether such hints can enhance LM performance (refer to Figure \ref{fig:prompt-task4-hint-example} in Appendix \ref{app:prompts}).  
\end{enumerate}  
The prompts are detailed in Appendix \ref{app:prompts}. 

\subsection{Evaluation Metrics}  

All models are evaluated using distinct metrics tailored to the nature of each task. For classification-related tasks, we employ accuracy (Acc.), precision (Prec.), recall (Rec.), and F1-score ($F_1$) to comprehensively assess the models' ability to correctly classify honorific levels. For translation tasks (Task 2 and Task 3), evaluation is conducted using BLEU~\cite{papineni-etal-2002-bleu} and CHRF++~\cite{popovic-2017-chrf}, as both metrics are well-suited for assessing machine translation (MT) quality. In CHRF++, the parameter $\beta$ is set to 2, as this configuration has been shown to achieve the highest Kendall’s $\tau$ segment-level correlation with human relative rankings (RR), ensuring more reliable evaluation results~\cite{popovic-2016-chrf}.

\section{Results and Analysis}
The experimental results for the fine-tuned models in Task 1 are presented in Table~\ref{tab:classifier-performance}. Additionally, the performance of the zero-shot classification task using Off-the-Shelf models is summarized in Table~\ref{tab:performance_classification_task}. The results of the honorific style translation task, evaluated using BLEU, are reported in Table~\ref{tab:performance_style_task_bleu}, while the complementary results based on the CHRF++ metric are provided in Table~\ref{tab:performance_style_task_chrf} in the Appendix. Similarly, the results of the honorific cross-lingual translation task, evaluated using BLEU, are detailed in Table~\ref{tab:performance_crosslingual_bleu}, with the corresponding CHRF++ results presented in Table~\ref{tab:performance_crosslingual_chrf} in the Appendix. Finally, the performance of the models on the conversation generation task is presented in Table~\ref{tab:performance_conversation_task} and further illustrated in Figure~\ref{fig:performance_radar}.  

\subsection{Baseline and Fine-tuned Honorific Level Classification}
\begin{table}[!th]
\centering
\resizebox{0.49\textwidth}{!}{
\begin{tabular}{lccccc}
\toprule
\textbf{Model} &  \textbf{Acc.} & \textbf{Prec.} & \textbf{Rec.} & \textbf{F1} \\
\midrule
\textbf{Baseline} \\
$\quad$Dictionary-Based Model & 88.37 & 88.92 & 88.37 & 88.64 \\
$\quad$LSTM Model & 93.47 & 91.56 & 92.78 & 91.34 \\ 
\midrule
\multicolumn{5}{l}{\textbf{Fine-tuning}} \\
$\quad$Javanese BERT & 93.91 & 94.09 & 93.91 & 93.97 \\
$\quad$Javanese GPT2 & 92.42 & 92.48 & 92.42 & 92.43 \\
$\quad$Javanese DistilBERT & \textbf{95.65} & \textbf{95.66} & \textbf{95.65} & \textbf{95.66} \\
\bottomrule
\end{tabular}
}
\caption{
Performance comparison of different fine-tuned models on Task 1: Honorific Level Classification.
}
\label{tab:classifier-performance}
\vspace{-1em}
\end{table}
The rule-based dictionary model achieved an accuracy of 88.37\%, as shown in Table~\ref{tab:classifier-performance}. While less flexible than machine learning approaches, this model provided a useful baseline for comparison. Additionally, the LSTM model, employed as a baseline deep learning approach, achieved an accuracy of 93.47\% (Table~\ref{tab:classifier-performance}). This result highlights the effectiveness of traditional neural networks in classifying Javanese honorific levels when trained on a well-annotated corpus. The strong performance of the LSTM model may be partially attributed to the relatively small dataset, as LSTM models tend to perform well with limited data~\citep{DBLP:journals/corr/abs-2009-05451}. 

Under identical experimental conditions, fine-tuned transformer-based models demonstrated superior performance compared to the baselines. \texttt{Javanese DistilBERT} achieved the highest accuracy at 95.65\%, surpassing both \texttt{Javanese BERT} (93.91\%) and \texttt{Javanese GPT-2} (92.42\%). 
The fine-tuned \texttt{Javanese DistilBERT} outperformed all other classifiers (Table~\ref{tab:classifier-performance} \& Table~\ref{tab:performance_classification_task}), making it the most suitable choice to automatically assess the honorific level of the generated text in Task 4. 

To assess robustness when classifier models are trained and tested on sentences with different meanings, Table \ref{tab:robustness_grouping} presents the results of cross-validation with grouping. We group sentences based on meaning before splitting them into folds. This ensures that sentences with the same meaning but different honorific levels are kept in the same fold, either all in the training set or all in the test set. As a result, models are trained to focus on differences in honorific levels rather than meaning correlations. The results demonstrate the model's strong generalization capabilities across different manners of splitting data.

\begin{table}[H]
\centering
\resizebox{0.49\textwidth}{!}{
\begin{tabular}{lcccc}
\toprule
\textbf{Model} & \textbf{Acc.} & \textbf{Prec.} & \textbf{Rec.} & \textbf{F1} \\
\midrule
\rowcolor{gray!20}\multicolumn{5}{c}{\textbf{Cross Validation}} \\
\midrule
Javanese BERT & \textbf{93.99} & \textbf{94.04} & \textbf{93.99} & \textbf{93.96} \\
Javanese GPT2 & 91.48 & 91.48 & 91.48 & 91.40 \\
Javanese DistilBERT & 93.10 & 93.14 & 93.10 & 93.05 \\
\bottomrule
\end{tabular}
}
\caption{Performance of fine-tuned models on Javanese honorific classification with cross-validation settings, where models are evaluated using 5-fold cross-validation on the training set to assess consistency and reliability.}
\label{tab:robustness_grouping}
\end{table}

\subsection{Off-the-Shelf Honorific Level Classification}

\begin{table}[h]
    \centering
    \resizebox{0.49\textwidth}{!}{
    \begin{tabular}{lcccc}
        \toprule
        \textbf{Model} & \textbf{Accuracy} & \textbf{Precision} & \textbf{Recall} & \textbf{F1} \\
        \midrule
        \multicolumn{5}{l}{\textbf{Closed-source}} \\
        $\quad$GPT-4o & \textbf{53.50} & 40.30 & 49.80 & 40.70 \\
        $\quad$Gemini 1.5 Pro & 50.70 & \textbf{54.60} & \textbf{54.20} & \textbf{45.40} \\
        \multicolumn{5}{l}{\textbf{Open-source}} \\
        $\quad$Sailor2 8B & 33.60 & 20.90 & 24.00 & 16.40 \\
        $\quad$Gemma2 9B Instruct & 28.40 & 42.30 & 23.00 & 17.80 \\
        $\quad$Llama3.1 8B Instruct & 43.00 & 20.60 & 30.50 & 24.00 \\
        $\quad$SahabatAI v1 Instruct (Llama3 8B) & 48.50 & 38.00 & 35.10 & 31.00 \\
        $\quad$SahabatAI v1 Instruct (Gemma2 9B) & 47.50 & 35.00 & 34.70 & 30.00 \\
        \midrule
        \rowcolor{gray!20} \multicolumn{5}{c}{\textbf{Honorific-Level Evaluation}} \\
        \midrule
        \textbf{Model} & \textbf{Level} & \textbf{Precision} & \textbf{Recall} & \textbf{F1} \\
        \midrule
        GPT-4o & Ngoko & \underline{78.00} & \textbf{\underline{91.10}} & \underline{84.00} \\
        & Ngoko Alus & 0 & 0 & 0 \\
        & Krama & 53.50 & 26.00 & 35.00 \\
        & Krama Alus & 29.90 & 82.40 & 43.80 \\
        \midrule
        Gemini 1.5 Pro & Ngoko & \textbf{\underline{90.10}} & 80.00 & \textbf{\underline{84.70}} \\
        & Ngoko Alus & 26.10 & 43.10 & 32.50 \\
        & Krama & 70.60 & 10.50 & 18.30 \\
        & Krama Alus & 31.70 & \underline{83.40} & 45.90 \\
        \midrule
        Sailor2 8B & Ngoko & 34.40 & \underline{86.30} & \underline{49.20} \\
        & Ngoko Alus & 4.00 & 0.30 & 0.6 \\
        & Krama & \underline{37.50} & 8.20 & 13.50 \\
        & Krama Alus & 7.50 & 1.30 & 2.30 \\
        \midrule
        Gemma2 9B Instruct & Ngoko & \underline{88.90} & 6.00 & 1.10 \\
        & Ngoko Alus & 5.60 & 10.50 & 7.30 \\
        & Krama & 36.80 & \underline{72.30} & \underline{48.80} \\
        & Krama Alus & 37.70 & 8.70 & 14.10 \\
        \midrule
        Llama3.1 8B Instruct & Ngoko & \underline{46.70} & \underline{87.70} & \underline{61.00} \\
        & Ngoko Alus & 0 & 0 & 0 \\
        & Krama & 35.80 & 34.40 & 35.10 \\
        & Krama Alus & 0 & 0 & 0 \\
        \midrule
        SahabatAI v1 Instruct  & Ngoko & \underline{61.20} & \underline{71.50} & \underline{65.90} \\
        (Llama3 8B) & Ngoko Alus & 19.40 & 3.40 & 5.80 \\
        & Krama & 40.60 & 64.20 & 49.70 \\
        & Krama Alus & 30.80 & 1.30 & 2.60 \\
        \midrule
        SahabatAI v1 Instruct  & Ngoko & \underline{51.80} & \underline{86.30} & \underline{64.70} \\
        (Gemma2 9B) & Ngoko Alus & 0 & 0 & 0 \\
        & Krama & 41.00 & 45.60 & 43.20 \\
        & Krama Alus & 47.20 & 7.00 & 12.20 \\
        \bottomrule
    \end{tabular}
    }
    \caption{Comparison of zero-shot overall and per-honorific level classification (Task 1) performance across different Off-the-Shelf models. The upper section reports overall Accuracy, Precision, Recall, and F1-score, while the lower section presents performance per honorific level. Additionally, detailed information about 0 result could be seen in Appendix.T~\ref{tab:performance_classification_task - percentage for zero result}.}
    \label{tab:performance_classification_task}
    \vspace{-1em}
\end{table}

In Table~\ref{tab:performance_classification_task}, among the closed-source and open-source models, \texttt{GPT-4o} achieved the highest accuracy (53.50\%) in Task 1: Honorific Level Classification. Meanwhile, \texttt{Gemini 1.5 Pro} obtained the highest scores in precision, recall, and F1-score.

An analysis of evaluation metrics per honorific label reveals a strong tendency among Off-the-Shelf models to classify texts into the \textit{Ngoko} label. This bias is particularly evident in \texttt{Gemini 1.5 Pro}, which achieved the highest precision (90.10\%) and F1-score (86.70\%) for the \textit{Ngoko} label, while \texttt{GPT-4o} attained the highest recall (91.10\%) for the same label. Additionally, the \textit{Ngoko} label consistently yielded the highest evaluation metric scores across nearly all Off-the-Shelf models, further reinforcing the classification bias toward this honorific level. Since these models perform poorly on other honorific levels, their overall performance remains lower than that of the fine-tuned models (Table~\ref{tab:classifier-performance}). This bias toward the \textit{Ngoko} honorific level may be attributed to the predominance of \textit{Ngoko} text in existing Javanese language corpora used to pre-train these LLMs.

\begin{table}[H]
    \centering
    \large
    \begin{adjustbox}{max width=\linewidth}
    \begin{tabular}{l|c|ccc}
        \toprule
         & & \multicolumn{3}{c}{\textbf{Misinterpreted Label}} \\
        \textbf{Model} & \textbf{Level} & \textbf{Ngoko (\%)} & \textbf{Krama (\%)} & \textbf{Krama Alus (\%)} \\
        \midrule
        GPT-4o & Ngoko Alus & 31.36 & 24.41 & 44.24 \\
        Llama3.1 8B Instruct & Ngoko Alus & 50.00 & 50.00 & 0 \\
        SahabatAI v1 Instruct & Ngoko Alus & 43.22 & 53.22 & 3.56 \\
        \midrule
        \textbf{Model} & \textbf{Level} & \textbf{Ngoko (\%)} & \textbf{Ngoko Alus (\%)} & \textbf{Krama (\%)} \\
        \midrule
        Llama3.1 8B Instruct & Krama Alus & 32.95 & 0 & 67.05 \\
        \bottomrule
    \end{tabular}
    \end{adjustbox}
    \caption{Percentage of label misclassification in the honorific level classification experiment where the evaluation metrics—Accuracy, Precision, and Recall—yield a value of 0.}
    \label{tab:performance_classification_task - percentage for zero result}
\end{table}

\begin{table*}[ht]
    \centering
    \Large  
    \begin{adjustbox}{max width=\textwidth}
    \begin{tabular}{lcccccccccccc}
        \toprule
        & \multicolumn{3}{c}{\textbf{Ngoko$\rightarrow$}X} & \multicolumn{3}{c}{\textbf{Ngoko Alus$\rightarrow$}X} & \multicolumn{3}{c}{\textbf{Krama$\rightarrow$}X} & \multicolumn{3}{c}{\textbf{Krama Alus$\rightarrow$}X} \\
        \cmidrule(lr){2-4} \cmidrule(lr){5-7} \cmidrule(lr){8-10} \cmidrule(lr){11-13}
        \textbf{Model} & \textbf{Ngoko} & \textbf{Krama} & \textbf{Krama} & 
        \textbf{Ngoko} & \textbf{Krama} & \textbf{Krama} &
        \textbf{Ngoko} & \textbf{Ngoko} & \textbf{Krama} &
        \textbf{Ngoko} & \textbf{Ngoko} & \textbf{Krama} \\
        & \textbf{Alus} & & \textbf{Alus} & 
        & & \textbf{Alus} &
        & \textbf{Alus} & \textbf{Alus} &   
        & \textbf{Alus} & \\
        \midrule
        \cellcolor{lightgray}Copy Baseline & \cellcolor{lightgray}30.33 & \cellcolor{lightgray}10.07 & \cellcolor{lightgray}8.23 & \cellcolor{lightgray}30.00 & \cellcolor{lightgray}8.66 & \cellcolor{lightgray}28.08 & \cellcolor{lightgray}10.00 & \cellcolor{lightgray}8.68 & \cellcolor{lightgray}31.46 & \cellcolor{lightgray}8.15 & \cellcolor{lightgray}28.07 & \cellcolor{lightgray}31.37 \\ 
        \midrule
        GPT-4o & 23.74 & \textbf{22.88} & \textbf{19.72} & \textbf{38.83} & \textbf{16.83} & \textbf{34.95} & \textbf{36.80} & \textbf{17.70} & \textbf{32.60} & \textbf{29.92} & 24.30 & 25.55 \\
        Gemini-1.5-pro & 2.52 & \textbf{11.77} & 4.36 & 24.71 & 5.97 & 5.28 & \textbf{27.39} & \textbf{12.94} & 0.78 & \textbf{21.73} & 15.49 & 9.36 \\
        Sailor2 8B & 1.02 & 1.13 & 1.15 & 1.01 & 1.04 & 0.89 & 1.19 & 0.90 & 0.40 & 1.24 & 1.12 & 0.59 \\
        Gemma2 9B Instruct & 0.73 & 0.21 & 0.14 & 1.01 & 0.33 & 0.83 & 0.31 & 0.22 & 0.76 & 0.20 & 0.79 & 1.22 \\
        Llama3.1 8B Instruct & 0.36 & 0.10 & 0.08 & 0.20 & 0.09 & 0.33 & 0.17 & 0.11 & 0.33 & 0.12 & 0.29 & 0.32 \\
        SahabatAI v1 (Llama3 8B) & 0.72 & 0.21 & 0.20 & 0.63 & 0.27 & 0.78 & 0.39 & 0.23 & 0.81 & 0.30 & 0.61 & 0.56 \\
        SahabatAI v1 (Gemma2 9B) & 0.47 & 0.23 & 0.39 & 0.69 & 0.13 & 0.34 & 0.61 & 0.36 & 0.35 & 0.47 & 0.49 & 0.29 \\
        \bottomrule
    \end{tabular}
    \end{adjustbox}
    \caption{Performance of honorific style translation across different models and  different honorific levels measured using BLEU. The \textbf{Copy Baseline} represents the BLEU score obtained by directly comparing the input text with the ground truth output, without any model inference. Bold text indicate model performance that surpass the Copy Baseline, indicating some style change capabilities.}
    \label{tab:performance_style_task_bleu}
    \vspace{-1em}
\end{table*}

Further analysis in Table~\ref{tab:performance_classification_task} and Table~\ref{tab:performance_classification_task - percentage for zero result} reveals that the \textit{Ngoko Alus} label is frequently misclassified, with errors spread across the other three labels. No instances are predicted as \textit{Ngoko Alus}, suggesting that models struggle to identify this honorific level. Misclassification is also observed for \textit{Krama Alus}, particularly in \texttt{LLaMA3.1 8B Instruct}, where it is often confused with \textit{Krama}, indicating difficulty in distinguishing between adjacent honorific levels.

\subsection{Honorific Style Change}

In this task, the \texttt{GPT-4o} model outperforms all other Off-the-Shelf models. Its performance surpasses some of the copy baseline, where inputs are directly compared to ground truth outputs without inference. This suggests that \texttt{GPT-4o} possesses some degree of style translation capability.  

Table~\ref{tab:performance_style_task_bleu} shows that when translation tasks are grouped by honorific level, the best results are observed in \textit{Ngoko} and \textit{Krama Alus}. Style translation toward these levels consistently exceeds the copy baseline, indicating that models perform better when distinguishing between honorific levels with significant differences, such as \textit{Ngoko} (the lowest level) and \textit{Krama Alus} (the highest). In contrast, the models struggle to differentiate between honorific levels that are closer in hierarchy, e.g., \textit{Ngoko} to \textit{Ngoko Alus} or \textit{Krama Alus} to \textit{Krama}.

\begin{figure*}[!th]
    \centering
    \small
    \includegraphics[width=0.85\textwidth]{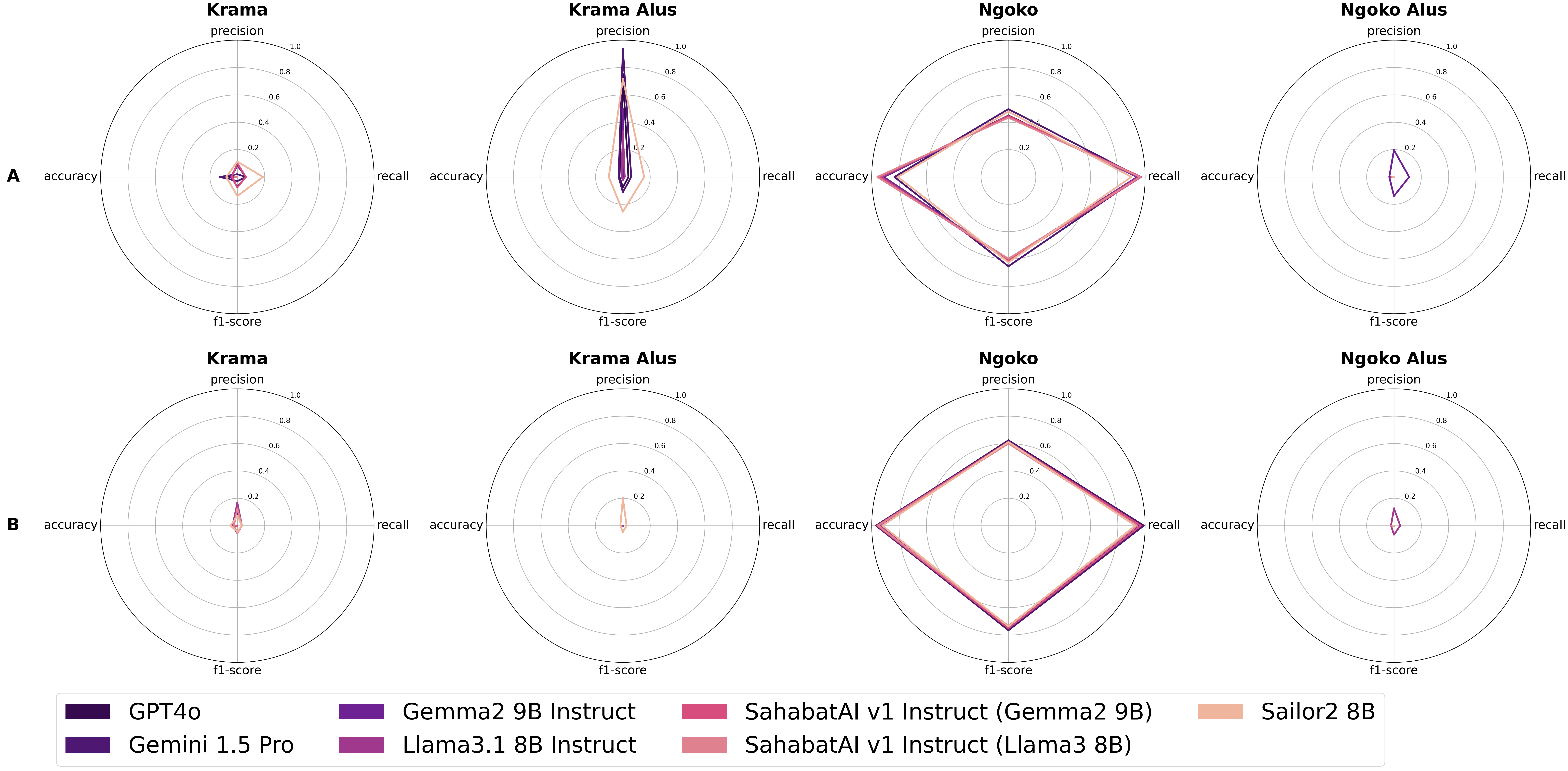}
    \caption{A comparative analysis of conversation generation quality across models, categorized by different honorific labels and the utterance (A or B).}
    \label{fig:performance_radar}
    \vspace{-1em}
\end{figure*}

\subsection{Cross-lingual Honorific Translation}

In this task, \texttt{GPT-4o} and \texttt{Gemini 1.5 Pro} are the top-performing models, while the others exhibit significantly lower performance. \texttt{GPT-4o} outperforms all models in nearly every translation category, except for \textit{Indonesian} $\rightarrow$ \textit{Krama}, where \texttt{Gemini 1.5 Pro} leads with a marginal BLEU score advantage of 0.55. In all other categories, \texttt{GPT-4o} achieves the highest performance, notably in \textit{Ngoko} $\rightarrow$ \textit{Indonesian} translation, where it attains 38.28 BLEU.

\begin{table}[htbp]
    \centering
    \large
    \resizebox{0.49\textwidth}{!}{
    \begin{tabular}{lcccc}
    \toprule
    & \multicolumn{4}{c}{\textbf{ID $\rightarrow$ X}} \\
    \cmidrule(lr){2-5}
    \multicolumn{1}{c}{\textbf{Model}} &
    \textbf{Ngoko} & 
    \textbf{Ngoko Alus} & 
    \textbf{Krama} & 
    \textbf{Krama Alus} \\
    \midrule
    GPT4o & \textbf{27.81} & \textbf{14.52} & 14.31 & \textbf{15.56} \\
    Gemini-1.5-pro & 16.76 & 6.99 & \textbf{ 14.86} & 7.05 \\
    Sailor2 8B & 4.1 & 1.2 & 0.89 & 0.98 \\
    Gemma2 9B Instruct & 0.43 & 0.29 & 0.12 & 0.11 \\
    Llama3.1 8B Instruct & 0.04 & 0.07 & 0.04 & 0.04 \\
    SahabatAI v1 Instruct (Llama3 8B) & 0.78 & 0.37 & 0.11 & 0.13 \\
    SahabatAI v1 Instruct (Gemma2 9B) & 6.27 & 3.49 & 1.06 & 1.59 \\
    \midrule
     & \textbf{Ngoko} & 
    \textbf{Ngoko Alus} & 
    \textbf{Krama} & 
    \textbf{Krama Alus} \\
    \cmidrule(lr){2-5}
    \multicolumn{1}{c}{\textbf{Model}} & \multicolumn{4}{c}{\textbf{X $\rightarrow$ ID}} \\
    \midrule
    GPT4o & \textbf{ 38.28} & \textbf{30.58} & \textbf{ 33.31} & \textbf{27.47} \\
    Gemini-1.5-pro & 32.74 & 23.82 & 29.34 & 23.81 \\
    Sailor2 8B & 7.27 & 6.36 & 6.71 & 5.8 \\
    Gemma2 9B Instruct & 3.29 & 2.22 & 1.55 & 1.77 \\
    Llama3.1 8B Instruct & 0.29 & 0.23 & 0.16 & 0.23 \\
    SahabatAI v1 Instruct (Llama3 8B) & 4.51 & 3.64 & 3.19 & 3.07 \\
    SahabatAI v1 Instruct (Gemma2 9B) & 15.0 & 11.5 & 10.39 & 9.27 \\
    \bottomrule
    \end{tabular}
    }
    \caption{Cross-lingual evaluation using BLEU scores for translation between Indonesian (ID) and Javanese honorific levels.}
    \label{tab:performance_crosslingual_bleu}
    \vspace{-0.5em}
\end{table}

As shown in Table~\ref{tab:performance_crosslingual_bleu}, translation \textbf{\textit{from Indonesian}} generally yields lower performance compared to translation \textbf{\textit{into Indonesian}}. For example, in the \textit{Indonesian} $\rightarrow$ \textit{Krama Alus} and \textit{Krama Alus} $\rightarrow$ \textit{Indonesian} categories, the \textit{Indonesian} $\rightarrow$ \textit{Krama Alus} translation achieves a BLEU score of only 15.56, whereas the \textit{Krama Alus} $\rightarrow$ \textit{Indonesian} translation yields a significantly higher BLEU score of 27.47. This finding confirms that translating from Javanese to Indonesian is generally easier for language models than the reverse.  

Among translations to Javanese, those targeting \textit{Ngoko} achieve the highest performance across models, mirroring the trend observed in Tasks 1 to 3, where LLMs exhibit a bias toward \textit{Ngoko}. 

\subsection{Conversation Generation}

\begin{table}[h]
    \centering
    \resizebox{0.49\textwidth}{!}{
    \begin{tabular}{lcccc}
        \toprule
        \multicolumn{1}{c}{} & \multicolumn{2}{c}{\textbf{Without Hint}} & \multicolumn{2}{c}{\textbf{With Hint}} \\
        \cmidrule(lr){2-3} \cmidrule(lr){4-5}
        \multicolumn{1}{c}{\textbf{Model}} & \textbf{A} & \textbf{B} & \textbf{A} & \textbf{B} \\
        \midrule
        GPT-4o & 43.12 & \textbf{ 60.00}  & 42.50 & 60.00 \\
        Gemini-1.5-pro & 46.25 & \textbf{ 60.00}  & 42.50 & \textbf{ 60.62}  \\
        Sailor2 8B & \textbf{ 48.75}  & 58.12 & \textbf{ 45.00}  & 56.25 \\
        Gemma2 9B Instruct & 42.50 & 58.12 & 41.25 & 58.12 \\
        Llama3.1 8B Instruct & 40.62 & 55.00 & 42.50 & 60.00 \\
        SahabatAI (Llama3 8B) & 41.88 & 57.50 & 41.25 & 57.50 \\
        SahabatAI (Gemma2 9B) & 41.25 & 58.75 & 43.12 & 58.75 \\
        \bottomrule
    \end{tabular}
    }
    \caption{Honorific level generation accuracy for conversation generation task.}
    \label{tab:performance_conversation_task}
    \vspace{-1em}
\end{table}

The \texttt{Sailor} model achieved the highest accuracy in classifying speaker \( A \)'s utterance (the first speaker) in both experiments, with and without hints (Table~\ref{tab:performance_conversation_task}). For speaker \( B \)'s utterance, the highest accuracy was attained by the \texttt{Gemini 1.5 Pro} and \texttt{GPT-4o} models. Providing hints in the form of usage guidelines for each honorific level did not consistently lead to  improvements.

Additionally, Figure~\ref{fig:performance_radar} presents the evaluation metrics for each honorific label, showing that the \textit{Ngoko} label consistently achieves the highest scores. These findings suggest that the models continue to struggle with fully understanding the nuanced usage of different honorific levels in conversation. Specifically, they tend to generate dialogues predominantly in \textit{Ngoko}, indicating a preference for this honorific over others.  

Notably, the models consistently perform better when generating utterances for speaker \( A \) than for speaker \( B \). As shown in Figure~\ref{fig:performance_radar}, speaker \( B \) surpasses speaker \( A \) only in the \textit{Ngoko} label. This suggests that models face greater difficulty in generating appropriate honorific-level speech when responding to a conversation, whereas they demonstrate greater proficiency in initiating dialogue with the correct honorific level.

\section{Conclusion and Future Work}

We introduce the Javanese Honorific Corpus ($\datasetname$) as a valuable linguistic resource and evaluate the ability of modern NLP models to handle Javanese honorific-related tasks, including classification, machine translation, and text generation. Our experimental results reveal that LLMs exhibit a bias toward a specific honorific level (\textit{Ngoko}), primarily due to the imbalanced distribution of honorific tiers in existing Javanese datasets.  

Additionally, cross-lingual performance on Javanese text with distinct honorific levels remains poor, and LLMs struggle to accurately interpret and generate honorifics in conversational contexts. This study’s methodologies and datasets provide a foundation for future research on low-resource languages with complex honorific systems. Future work may explore dialectal variations, finer linguistic nuances, or larger datasets to enhance model performance and generalizability.

\section*{Limitations}
\paragraph{Corpus Size}
The $\datasetname$ contains 4,024 sentences and 160 conversation cases, which may not capture the full complexity of Javanese honorific. Expanding the corpus with more diverse sources, including spoken language, would improve generalizability.

\paragraph{Dialect}
The corpus does not account for regional dialects of Javanese so it limits model accuracy in different linguistic regions. Future work should include dialectal variations for broader coverage.

\section*{Ethical Consideration}
Misclassification of honorifics can result in disrespectful interactions within Javanese social contexts. It is crucial for users of these models to be aware of their limitations in handling language features that are culturally sensitive.

\section*{Acknowledgements}
This research was funded by Lembaga Pengelola Dana Pendidikan (LPDP), the Ministry of Education, Culture, Research and Technology of the Republic of Indonesia through the Indonesia-US Research Collaboration in Open Digital Technology program, and Monash University's Action Lab. Their support for this research is deeply appreciated, and we acknowledge their vital role in the successful completion of this work. The findings and conclusions presented in this publication are those of the authors and do not necessarily reflect the views of the sponsors.

\bibliographystyle{acl_natbib} 
\bibliography{custom}          

\clearpage
\appendix

\begin{table*}[!th]
    \renewcommand{\arraystretch}{1.3}
    \centering
    \resizebox{\textwidth}{!}{
    \begin{tabular}{lcccccc}
        \toprule
        \textbf{Honorific Level} & \multicolumn{3}{c}{\textbf{Affixes (Tataran Rimbag)}} & \multicolumn{3}{c}{\textbf{Word Type Levels (Tataran Tembung)}} \\
        & Krama Inggil & Krama & Ngoko & Krama Inggil & Krama & Ngoko \\ 
        \midrule
        Krama Alus & \cmark & \cmark & \xmark & \cmark & \cmark & \cmark \\
        Krama & \xmark & \cmark & \xmark & \xmark & \cmark & \cmark \\
        Ngoko Alus & \cmark & \xmark & \cmark & \cmark & \xmark & \cmark \\
        Ngoko & \xmark & \xmark & \cmark & \xmark & \xmark & \cmark \\
        \bottomrule
    \end{tabular}
    }
    \caption{The mapping of Honorific Levels includes their corresponding affixes and word type levels.}
    \label{tab:system}
    \renewcommand{\arraystretch}{1.0}
\end{table*}

\label{app:unggahungguhbasa}
\section{Unggah-Ungguh Basa System}
\begin{figure*}[!ht]
    \centering
    \small
    \includegraphics[width=0.98\textwidth]{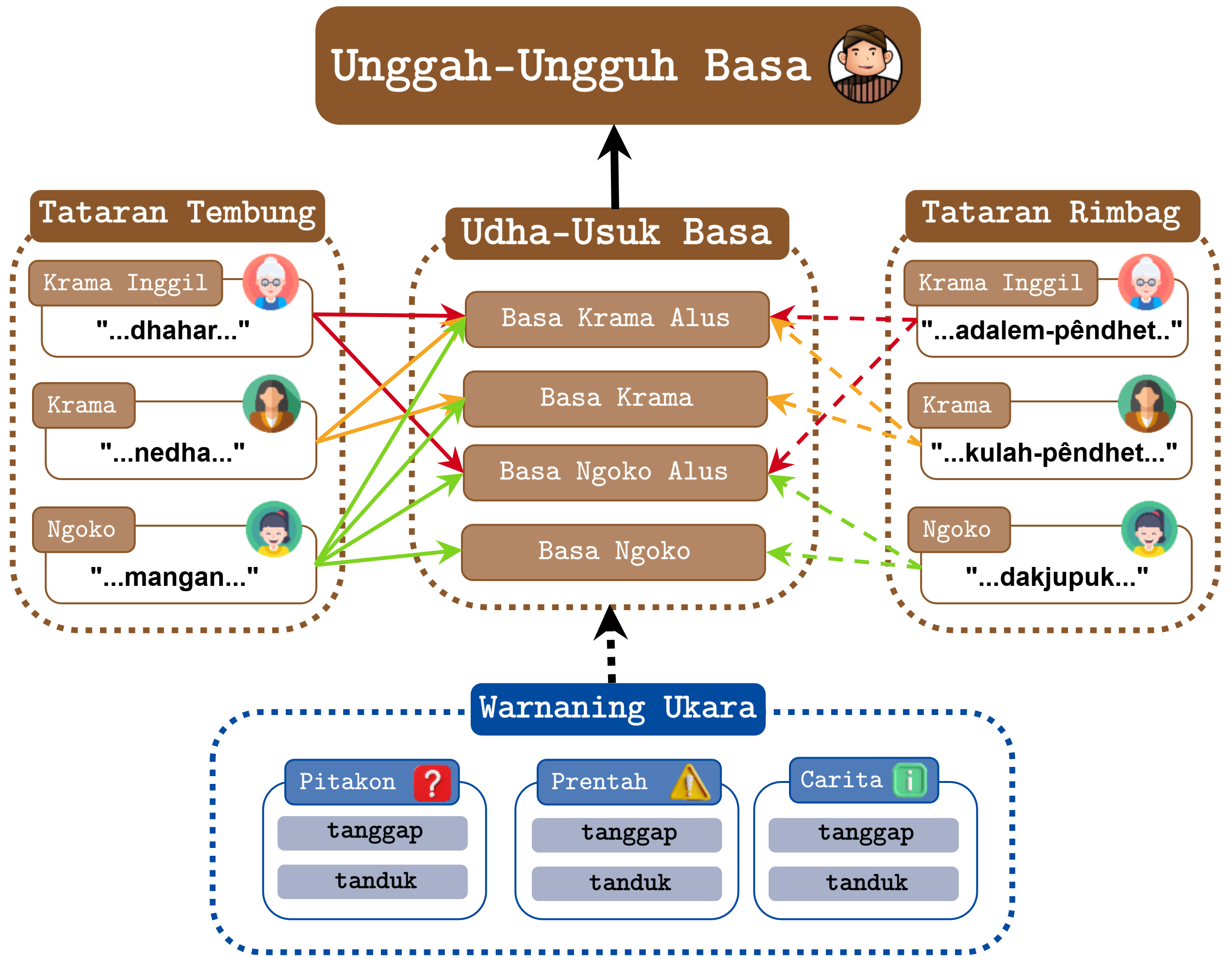}  
    \caption{Diagram of the Unggah-Ungguh Basa System.}
    \label{fig:diagram}
\end{figure*}

The Unggah-Ungguh Basa system (Figure~\ref{fig:diagram}) is a Javanese linguistic framework designed to guide the appropriate use of varying honorific levels depending on conversational context. Based on the work of \citet{Harjawiyana2009} and detailed in Table~\ref{tab:system}, this system comprises four main components: Word Type Levels (Tataran Tembung), Affixes (Tataran Rimbag), and Sentence Type (Warnaning Ukara).

\paragraph{Word Type Levels (Tataran Tembung).}
This categorization is based on speech formality, ranging from informal (Ngoko) to polite (Krama), and extending to the most formal expressions (Krama Inggil).

\paragraph{Affixes (Tataran Rimbag).}
This process involves modifying the root of a word by adding a prefix, infix, or suffix, and encompasses three distinct honorific affixes:
\begin{itemize}
    \item Less formal affixes (Rimbag Ngoko): ``dakjupuk'', ``kokjupuk'', ``dijupukake''.
    \item Formal affixes (Rimbag Krama): ``kulah-pêndhet'', ``sampêyan-pêndhet'', ``dipundhêtaken''.
    \item The most formal affixes (Rimbag Krama Inggil): ``adaleṃ-pêndhet'', ``panjenengan-pundhut''.
\end{itemize}

\paragraph{Sentence Type (Warnaning Ukara).}
This system encompasses various sentence types, such as commands, requests, narratives, and questions. Responses can be either "tanggap" (active) or "tanduk" (passive).

\paragraph{Language Hierarchical Structure (Undha-Usuk Basa).}
This represents the hierarchical structure of social interaction or communication as follows:
\begin{itemize}
    \item Ngoko: Emerges from informal (Ngoko) speech combined with informal discourse (Rimbag ngoko).
    \item Ngoko Alus: Combines informal (Ngoko) words with high or polite speech levels to reflect politeness.
    \item Krama: Involves polite (Krama) discourse combined with either informal or polite words, and speech levels of Krama or Ngoko.
    \item Krama Alus: Incorporates formal speech with polite words, indicating a high level of politeness.
\end{itemize}
This system highlights the complex interplay of formality and social interaction in Javanese language.

\section{Dataset}
\justifying

\subsection{$\datasetname$'s Samples}
\label{app:unggahungguhsamples}
The $\datasetname$ dataset comprises four columns: label, Javanese sentence, group, and Indonesian sentence (Table~\ref{tab:unggahungguh_samples}). The label column represents the \textbf{honorific level of each Javanese sentence} instance, where label "0" corresponds to the \textit{ngoko} honorific level, label "1" to \textit{ngoko alus}, label "2" to \textit{krama}, and label "3" to \textit{krama alus}. Sentences with the same meaning are grouped under a single group ID, as specified in the "group" column.
\begin{table*}[h]
    \centering
    \resizebox{0.98\textwidth}{!}{
    \begin{tabular}{c|p{6cm}|c|p{6cm}|p{6cm}}
        \toprule
        \textbf{Label} & \textbf{Javanese Sentence} & \textbf{Group} & \textbf{Indonesian Sentence} & \textbf{English Sentence} \\
        \midrule
        0 & Nggaawa jeruk kuwi! & 1 & Membawalah jeruk itu. & Bring that orange. \\
        1 & Panjenengan ngampil jeruk kuwi! &  &  \\
        2 & Sampeyan mbekta jeram menika! &  &  \\
        3 & Panjenengan ngampil jeram menika! &  &  \\
        \midrule
        0 & Tustele rusak, digawa ya ora ana gunane. & 2 & Tustelnya rusak, seandainya dibawa tidak ada gunanya. & The camera is broken, so there's no point in bringing it. \\
        1 & Tustele rusak, dipamila ya ora ana gunane. &  &  \\
        2 & Tustelipun risak, dipunbektaa inggih boten wonten ginanipun. &  &  \\
        3 & Tustelipun risak, dipunamila inggih boten wonten ginanipun. &  &  \\
        \midrule
        0 & Mobile dak-undurake, abanana! & 3 & Mobil saya undurkan, berilah aba-aba! & I am reversing my car, give me a signal! \\
        1 & Mobile adalem undurake, panjenengan paringi dhawu! &  &  \\
        2 & Mobilipun kula unduraken, sampeyan abani! &  &  \\
        3 & Mobilipun adalem unduraken, panjenengan paringi dhawu! &  &  \\
        \midrule
        0 & Yen wedange arep diladekake, ngabanana! & 4 & Jika minuman akan disajikan, memberilah isyarat! & If the drinks are going to be served, give a signal! \\
        1 & Yen unjukane arep diladekake, panjenengan mringi dhawu! &  &  \\
        2 & Yen wedangipun badhe dipunladosaken, sampeyan ngabani! &  &  \\
        3 & Yen unjukanipun badhe dipunladosaken, panjenengan maringi dhawu! &  &  \\
        \bottomrule
    \end{tabular}
    }
    \caption{Samples from $\datasetname$ Dataset.}
    \label{tab:unggahungguh_samples}
\end{table*}

\subsection{Conversation's Role and Context}
\label{app:conversationrole}
\begin{table}[H]
    \centering
    \small
    \resizebox{0.49\textwidth}{!}{
    \begin{tabular}{l|c|c|c}
        \toprule
        \textbf{Honorific level} &
        \textbf{A Utterance} &
        \textbf{B Utterance} &
        \textbf{Total} \\
        \midrule
        \midrule
        Ngoko & 69 & 97 & 166 \\
        Ngoko alus & 9 & 11 & 20 \\
        Krama & 8 & 15 & 23 \\
        Krama alus & 74 & 37 & 111 \\
        \bottomrule
    \end{tabular}
    }
    \caption{Honorific level distribution of conversation's status and context dataset.}
    \label{tab:Conversation Status and Context (Honorific Distribution)}
\end{table}
The \textit{Conversation’s Role and Context} dataset contains \textbf{160 instances} and is utilized for evaluating language models (LMs) in the task of \textbf{conversation generation} involving two speakers (A and B). It comprises \textbf{5 primary columns}, which are used to assess the model's ability to perform the task, along with \textbf{2 additional columns} that provide example dialogues for both speakers. The \textbf{primary columns} include: role A, role B, context, and honorific level for Speaker A's utterance \& B's utterance.  \textbf{Role A, Role B, and Context} columns are written in \textbf{English} to enhance accuracy when used in English-language prompts during LM inference. Meanwhile, the distribution of honorific level instances for \textbf{Speaker A's} and \textbf{Speaker B's utterance} columns can be seen in Table~\ref{tab:Conversation Status and Context (Honorific Distribution)}. Samples of the \textit{Conversation's Role and Context} primarily column dataset are presented in Table~\ref{tab:Conversation Status and Context}. These evaluation scenarios will be released publicly along with the \datasetname ~dataset. 

\begin{table*}[!h]
    \centering
    \Large
    \resizebox{0.98\textwidth}{!}{
    \begin{tabular}{l|l|p{5cm}|c|c}
        \toprule
        \textbf{Role A} &
        \textbf{Role B} &
        \textbf{Context} &
        \textbf{A Utterance Category} &
        \textbf{B Utterance Category} \\
        \midrule
        Teacher & Student & A asked speaker B about what he had eaten today.  Speaker A has a higher status or position than Speaker B. & 0 & 2 \\
        \midrule
        Peer & Peer & Speaker A asks speaker B about what Speaker B's father ate today. Speaker A and Speaker B have equal status or position and have familiar interactions. & 1 & 1 \\
        \midrule
        Older sibling &Younger sibling &Speaker A asks speaker B about what Speaker B learned at school today. Speaker B has a lower status or position than Speaker A. & 0 & 1 \\
        \midrule
        Grandchild & Grandmother & Speaker A gives him speaker B's glasses. Speaker A has a lower status or position than Speaker B. & 3 & 0 \\
        \bottomrule
    \end{tabular}
    }
    \caption{Primary column samples of conversation's status and context dataset.}
    \label{tab:Conversation Status and Context}
\end{table*}

\section{Prompt Template}
\label{app:prompts}
We utilize \textbf{6 distinct straightforward prompts} (Fig.~\ref{fig:prompt-task1-example}, \ref{fig:prompt-task2-example}, \ref{fig:prompt-task3-example}, \ref{fig:prompt-task4-nohint-example}, and \ref{fig:prompt-task4-hint-example}), each designed for a specific objective, covering four tasks along with several variations of certain tasks. The prompts used for the \textit{honorific style change} and \textit{honorific cross-lingual translation} tasks share a dominant structural similarity, with only minor modifications. Meanwhile, the \textbf{hints} included in the \textit{conversation generation} prompts serve as guidelines for the appropriate usage of different levels of Javanese honorific. These hints are intended to facilitate the conversation generation process, as the context provided to the model includes information extracted from the hints, particularly regarding the relationship between role A and role B. It is important to note that text highlighted in \textcolor{red}{red} within the prompts represents input that needs to be provided.

\section{Generation Result}
\subsection{Honorific Style Change}
Table~\ref{tab:Generation Sample - Style} presents the inference results of the Off-the-Shelf models on the honorific style change task. The GPT-4o and Gemini 1.5 Pro models produce outputs that are relatively comparable to the ground truth for each instance. In contrast, the other five models exhibit significantly lower performance. The Sailor2 8B model, when tested on instances from the fourth group of the $\datasetname$ dataset, generates outputs that are nearly correct but contain subordinate clauses that can be identified as hallucinations, as they deviate from the given input context. Meanwhile, the Gemma2 9B Instruct and SahabatAI v1 Instruct (Llama3 8B) models show a similar tendency of failing to modify the input appropriately, often producing translations that include vocabulary outside the Javanese language. Additionally, the Llama3.1 8B Instruct and SahabatAI v1 Instruct (Llama3 8B) models exhibit a comparable pattern, frequently translating the given input into Indonesian rather than transforming it into the intended honorific level.  

\begin{table*}[h]
    \centering
    \resizebox{0.98\textwidth}{!}{
    \begin{tabular}{l|p{5cm}|p{5cm}|p{5cm}}
        \toprule
        \textbf{Model} & \textbf{Input (\textit{Ngoko})} & \textbf{Generation} & \textbf{Gold (\textit{Krama Alus})} \\
        \midrule
        GPT 4o & Rambutmu ambune kok wangi, nganggo sampo apa? & Rambut panjenengan ambune kok wangi, nganggo sampo punapa? & Rikma panjenengan gandanipun kok wangi, ngagem sampo menapa? \\
        \midrule
        Gemini 1.5 Pro & Bu Haiti arep tuku gula selawe kilo, wis dak-dolane. & Ibu Haiti badhé mundhut gula kalih dasa gangsal kilo, sampun kula damelaken. & Bu Haiti badhe mundhut gendhis selangkung kilo, sampun adalem sadeanipun. \\
        \midrule
        Sailor2 8B & Yen wedange arep diladekake, ngabanana! & Yen wedangipun badhe dipunladosaken, kula aturi dhateng ing papan. & Yen unjukanipun badhe dipunladosaken, panjenengan maringi dhawuh! \\
        \midrule
        Gemma2 9B Instruct& Adhimu iku alise nanggal sapisan. & Nubarkah iku alise nanggal sapisan. & Ingkang rayi menika imbanipun nanggal sapisan. \\
        \midrule
        Llama3.1 8B Instruct & Aku arep matur ana ing ngarepmu kabeh. & Saya harap sekali dapat menyapa kawan-kawan seluruhnya. & Adalem badhe munjuk wonten ing ngarsa panjenengan sadaya. \\
        \midrule
        SahabatAI v1 Instruct (Llama3 8B) & Para kawula njaluk supaya Raden Wijaya madeg ratu ing Majapait. & Para satriya ngajak agar Prabhu Jayanagara dadi raja ing Majapahit. & Para kawula nyuwun supados Raden Wijaya jumeneng ratu ing Majapait. \\
        \midrule
        SahabatAI v1 Instruct (Gemma2 9B) & Mobile dak-undurne, abanana! & `Mobile`ipun ngaktipaken banjur dipununduh. & Mobilipun adalem unduraken, panjenengan paringi dhawuh! \\
        \bottomrule
    \end{tabular}
    }
    \caption{Generation samples of honorific style change task (\textit{Ngoko}$\rightarrow$\textit{Krama Alus}).}
    \label{tab:Generation Sample - Style}
\end{table*}

\subsection{Honorific Cross-lingual Translation}
\label{app:translationsamples}
The inference results for the honorific cross-lingual translation task, presented in Tables~\ref{tab:Generation Sample (1) - Crosslingual} \&~\ref{tab:Generation Sample (2) - Crosslingual}, indicate that translation performance for the "\textbf{to Indonesia}" direction is generally better than for the "\textbf{from Indonesia}" direction. The GPT-4o and Gemini 1.5 Pro models continue to demonstrate superior performance, while the other models tend to produce lower-quality outputs. These models frequently exhibit hallucinations, generating translations that are entirely unrelated to the input text. Additionally, they often misinterpret certain words, including common vocabulary. For instance, the SahabatAI v1 Instruct (Gemma2 9B) model incorrectly translates \textit{jeruk} (orange) as \textit{jambu} (guava).

\begin{table*}[h]
    \centering
    \resizebox{0.98\textwidth}{!}{
    \begin{tabular}{l|p{5cm}|p{5cm}|p{5cm}}
        \toprule
        \textbf{Model} & \textbf{Input (\textit{Ngoko})} & \textbf{Generation} & \textbf{Gold (\textit{Indonesian})} \\
        \midrule
        GPT 4o & Abang-abang kae klambi apa serbet? & Merah-merah itu baju apa serbet? & Merah-merah itu baju atau serbet? \\
        \midrule
        Gemini 1.5 Pro & Menakjinggane kae, sing raine diabang-abang. & Hebat sekali dia, yang mukanya merah padam. & Menakjingganya itu, yang wajahnya diberi warna merah! \\
        Sailor2 8B & Mobile dak-undurne, abanana! & Ia tidak mundur dari yang sudah-sudah, & Mobil saya undurkan, berilah aba-aba! \\
        \midrule
        Gemma2 9B Instruct& Abot-abot nggawa tela iku arep dianggo apa? & Apa yang akan dibawa oleh semua orang itu untuk digunakan? & Berat-berat membawa ketela itu akan dipergunakan untuk apa? \\
        \midrule
        Llama3.1 8B Instruct & Koran-koran sing wis ora kanggo iku dak-dolane wae. & Terakhir sekali koran itu tidak berguna. & Koran-koran yang sudah tidak terpakai itu saya jualinya saja. \\
        \midrule
        SahabatAI v1 Instruct (Llama3 8B) & Tustele rusak, digawaa ya ora ana gunane. & Kerusakan terjadi pada bagian yang tidak berguna. & Tustelnya rusak, seandainya dibawa tidak ada gunanya. \\
        \midrule
        SahabatAI v1 Instruct (Gemma2 9B) & Nggawaa jeruk kuwi! & Ambil jeruk itu! & Membawalah jeruk itu. \\
        \bottomrule
    \end{tabular}
    }
    \caption{Generation samples of honorific cross-lingual translation task (\textit{Ngoko}$\rightarrow$\textit{Indonesian}).}
    \label{tab:Generation Sample (1) - Crosslingual}
\end{table*}

\begin{table*}[!h]
    \centering
    \resizebox{0.98\textwidth}{!}{
    \begin{tabular}{l|p{5cm}|p{5cm}|p{5cm}}
        \toprule
        \textbf{Model} & \textbf{Input (\textit{Indonesian})} & \textbf{Generation} & \textbf{Gold (\textit{Ngoko})} \\
        \midrule
        GPT 4o & Merah-merah itu baju atau serbet? & Abang-abang kuwi klambi apa serbet? &  Abang-abang kae klambi apa serbet? \\
        \midrule
        Gemini 1.5 Pro & Menakjingganya itu, yang wajahnya diberi warna merah! & Nggilani kuwi, sing raine diwenehi werna abang! & Menakjinggane kae, sing raine diabang-abang. \\
        \midrule
        Sailor2 8B & Mobil saya undurkan, berilah aba-aba! & Mobile kulo munduri., & Mobile dak-undurne, abanana! \\
        \midrule
        Gemma2 9B Instruct & Berat-berat membawa ketela itu akan dipergunakan untuk apa? &  kono ketale iki akan dipergunakan untuk apa & Abot-abot nggawa tela iku arep dianggo apa? \\
        \midrule
        Llama3.1 8B Instruct & Koran-koran yang sudah tidak terpakai itu saya jualinya saja. & wong-wong wis ora, iki aku tuku. & Koran-koran sing wis ora kanggo iku dak-dolane wae. \\
        \midrule
        SahabatAI v1 Instruct (Llama3 8B) & Tustelnya rusak, seandainya dibawa tidak ada gunanya. & Palingan kanggo nggemborke wong-wong neng omah. & Tustele rusak, digawaa ya ora ana gunane. \\
        \midrule
        SahabatAI v1 Instruct (Gemma2 9B) & Membawalah jeruk itu. & Gawawo woh jambu iku! & Nggawaa jeruk kuwi! \\
        \bottomrule
    \end{tabular}
    }
    \caption{Generation samples of honorific cross-lingual translation task (\textit{Indonesian}$\rightarrow$\textit{Ngoko}).}
    \label{tab:Generation Sample (2) - Crosslingual}
\end{table*}

\subsection{Conversation Generation}
\label{app:convgenerationsamples}
Table~\ref{tab:Generation Sample - Conversation Generation} presents the generated outputs from the conversation generation task for each Off-the-Shelf model using specific instances. In general, the inference results across all models exhibit errors, particularly the inclusion of Indonesian vocabulary within the generated conversational sentences. Commonly misused words (Indonesian vocabulary) include apa kabar (how are you), saya (I), dimakan (eaten), mangga (mango), nasi (rice), aku (I), baru (new), and disini (here). Additionally, grammatical errors are observed in the model outputs. For instance, the inference result from the Gemini 1.5 Pro model contains a sub-clause that reads, "Iyo, sing sinau sregep...", which has an incorrect grammatical structure. The correct phrasing should be "Iyo, sinau sing sregep...".

\begin{table*}[h]
    \centering
    \resizebox{0.98\textwidth}{!}{
    \begin{tabular}{p{3cm}|p{4cm}|p{4cm}|p{4cm}|p{4cm}|p{4cm}}
    \hline
    \textbf{Model} & \textbf{Role A} & \textbf{Role B} & \textbf{Context} & \textbf{A Utterance} & \textbf{B Utterance} \\
    \hline
    GPT 4o & Schoolmate & Schoolmate & The speaker talked with his schoolmates at a reunion event. Speaker A and Speaker B have equal status or position and have familiar interactions. & Aku uga apik. Kangen karo jaman sekolah mbiyen, ya? & Iyo, kangen banget. Kowe isih eling wektu kita sering dolanan bareng sak kelas? \\
    \midrule
    Gemini 1.5 Pro & Grandfather & Grandchild & Speaker A gives pocket money to speaker B to provide for going to school. Speaker A has a higher status or position than Speaker B. & Iyo, sing sinau sregep. Aja dolan wae. (Yes, study diligently. Don't just play) & Inggih, Mbah. Kula badhe sregep sinau \\
    \midrule
    Sailor2 8B & A friend from the same village who has not met each other for a long time & A friend from the same village who has not met each other for a long time & Speaker A asks speaker B about the news from speaker B. Speaker A and Speaker B have equal status or position and have familiar interactions. & Hai, apa kabar? Wis suwe aku ora ketemu kowe. & Kabarku apik-apik wae Mas. Sampeyan piye? \\
    \midrule
    Gemma2 9B Instruct & People who do not know each other & People who do not know each other & Speaker A asked speaker B about the direction to Gadjah Mada University. Speaker A has a higher status or position than Speaker B & Mbak, ngerti ora mboten arah kanggo Universitas Gadjah Mada? & Maaf Pak, aku kurang tau kok. Aku baru di sini. \\
    \midrule
    Llama3.1 8B Instruct & Older sibling & Younger sibling & Speaker A asks speaker B about what Speaker B learned at school today. Speaker B has a lower status or position than Speaker A. & Saya kudu tahu apa kang wedhe siswamu sabene dina, ndak? & Mbok, saya iki wis sinaos nulis pethi ing buku lan nyimpening ngisor majelise. \\
    \midrule
    SahabatAI v1 Instruct (Llama3 8B) & Peer & Peer & Speaker A asks speaker B about what Speaker B's father ate today. Speaker A and Speaker B have equal status or position and have familiar interactions. & Apa sing dimakan bapakmu dina iki? & Mbok ya nasi goreng, tapi aku ora ngerti pasti \\
    \midrule
    SahabatAI v1 Instruct (Gemma2 9B) & Teacher & Student & Speaker A asked speaker B about what he had eaten today. Speaker A has a higher status or position than Speaker B. & Nggih, nak? Sampeyan wis mangan apa dina iki? & Mangga lan roti goreng \\
    \hline
    \end{tabular}
    }
    \caption{Samples of conversation generation task (A and B utterance).}
    \label{tab:Generation Sample - Conversation Generation}
\end{table*}

\section{Additional Results}

\subsection{Honorific Style Change}
\label{app:resultstyle}
Table~\ref{tab:performance_style_task_chrf} presents the performance evaluation of various models in translating between different Javanese honorific levels, measured using CHRF++ scores. The Copy Baseline represents the CHRF++ score obtained by directly comparing the input text with the ground truth without model inference, serving as a reference point for assessing model effectiveness. The results indicate that GPT-4o achieves the highest performance, surpassing the baseline across all translation tasks. Gemini-1.5-pro also demonstrates notable performance, particularly in Ngoko $\rightarrow$ Krama and Krama $\rightarrow$ Krama Alus translations. In contrast, models such as Sailor2 8B, Gemma2 9B Instruct, and Llama3.1 8B Instruct exhibit significantly lower scores, suggesting limited effectiveness in handling honorific style transformations. Table~\ref{tab:translation_instances - style} provides the distribution of sentence instances for each honorific level translation pair. The dataset is balanced across the different transformations, with the highest number of instances (1,412) observed in the Ngoko $\leftrightarrow$ Krama pair. This distribution ensures that models have sufficient training data for evaluating performance across both direct and reverse translations. However, the lower number of instances for Ngoko Alus transformations may contribute to the models' difficulty in accurately translating this level.

\begin{table*}[h]
    \centering
    \Large  
    \begin{adjustbox}{max width=\textwidth}
    \begin{tabular}{lcccccccccccc}
        \toprule
        & \multicolumn{3}{c}{\textbf{Ngoko$\rightarrow$}X} & \multicolumn{3}{c}{\textbf{Ngoko Alus$\rightarrow$}X} & \multicolumn{3}{c}{\textbf{Krama$\rightarrow$}X} & \multicolumn{3}{c}{\textbf{Krama Alus$\rightarrow$}X} \\
        \cmidrule(lr){2-4} \cmidrule(lr){5-7} \cmidrule(lr){8-10} \cmidrule(lr){11-13}
        \textbf{Model} & \textbf{Ngoko} & \textbf{Krama} & \textbf{Krama} & 
        \textbf{Ngoko} & \textbf{Krama} & \textbf{Krama} &
        \textbf{Ngoko} & \textbf{Ngoko} & \textbf{Krama} &
        \textbf{Ngoko} & \textbf{Ngoko} & \textbf{Krama} \\
        & \textbf{Alus} & & \textbf{Alus} & 
        & & \textbf{Alus} &
        & \textbf{Alus} & \textbf{Alus} &   
        & \textbf{Alus} & \\
        \midrule
        \cellcolor{lightgray}Copy Baseline & \cellcolor{lightgray}50.16 & \cellcolor{lightgray}34.71 & \cellcolor{lightgray}28.63 & \cellcolor{lightgray}54.69 & \cellcolor{lightgray}31.58 & \cellcolor{lightgray}59.61 & \cellcolor{lightgray}38.81 & \cellcolor{lightgray}32.48 & \cellcolor{lightgray}56.55 & \cellcolor{lightgray}33.12 & \cellcolor{lightgray}63.35 & \cellcolor{lightgray}58.36 \\ 
        \midrule
        GPT-4o & 47.00 & \textbf{52.66} & \textbf{49.02} & \textbf{59.79} & \textbf{45.34} & \textbf{65.28} & \textbf{60.80} & \textbf{39.51} & \textbf{60.32} & \textbf{53.03} & 48.21 & 52.89 \\
        Gemini-1.5-pro & 24.14 & \textbf{41.89} & 22.47 & 49.28 & \textbf{32.55} & 30.07 & \textbf{51.88} & \textbf{36.36} & 16.05 & \textbf{46.01} & 41.55 & 37.35 \\
        Sailor2 8B & 19.57 & 20.17 & 21.10 & 18.68 & 19.75 & 21.72 & 20.72 & 19.42 & 18.97 & 20.54 & 20.22 & 17.79 \\
        Gemma2 9B Instruct & 14.23 & 12.82 & 11.99 & 14.44 & 12.61 & 15.40 & 12.56 & 12.62 & 15.51 & 11.79 & 15.73 & 16.49 \\
        Llama3.1 8B Instruct & 13.26 & 13.15 & 12.61 & 11.88 & 12.96 & 15.81 & 12.31 & 12.31 & 14.42 & 11.28 & 15.38 & 14.26 \\
        SahabatAI v1 (Llama3 8B) & 15.86 & 14.72 & 14.62 & 13.92 & 14.80 & 18.93 & 14.98 & 14.05 & 17.27 & 14.11 & 17.12 & 15.83 \\
        SahabatAI v1 (Gemma2 9B) & 17.19 & 16.29 & 17.90 & 16.88 & 15.42 & 18.97 & 17.85 & 15.91 & 18.12 & 16.25 & 18.38 & 16.18 \\
        \bottomrule
    \end{tabular}
    \end{adjustbox}
    \caption{Performance evaluation of style translation models across different Javanese honorific levels, measured using CHRF++ scores. The \textbf{Copy Baseline} represents the CHRF++ score obtained by directly comparing the input translation text with the ground truth, without undergoing model inference. Bold text indicate model performance surpassing the Copy Baseline, demonstrating effective translation capabilities.}
    \label{tab:performance_style_task_chrf}
\end{table*}

\begin{table}[h]
    \centering
    \resizebox{0.49\textwidth}{!}{
    \begin{tabular}{ll|c}
        \toprule
        \multicolumn{2}{c|}{Translation Pairs (Bidirectional)} & Number of Instances \\
        Label 1 & Label 2 & (Sentences) \\
        \midrule
        \midrule
        Ngoko & Ngoko Alus & 585 \\
        Ngoko & Krama & 1412 \\
        Ngoko & Krama Alus & 595 \\
        Ngoko Alus & Krama & 584 \\
        Ngoko Alus & Krama Alus & 588 \\
        Krama & Krama Alus & 591 \\
        \bottomrule
    \end{tabular}
    }
    \caption{Number of sentence instances for each honorific level translation process. Each translation process is bidirectional, meaning the count for a given pair (e.g., Ngoko $\leftrightarrow$ Krama) is identical in both directions.}
    \label{tab:translation_instances - style}
\end{table}

\subsection{Honorific Cross-lingual Translation}
\label{app:resulttranslation}
Table~\ref{tab:performance_crosslingual_chrf} presents the CHRF++ scores for evaluating the cross-lingual translation between Indonesian (ID) and different Javanese honorific levels. The results indicate that GPT-4o consistently achieves the highest scores across all translation directions, demonstrating superior performance in both ID$\rightarrow$Javanese and Javanese$\rightarrow$ID tasks. Gemini-1.5-pro follows closely, particularly excelling in ID$\rightarrow$Krama and Krama$\rightarrow$ID translations. In contrast, models such as Sailor2 8B, Gemma2 9B Instruct, and Llama3.1 8B Instruct exhibit substantially lower scores, suggesting challenges in handling both Indonesian-to-Javanese and Javanese-to-Indonesian honorific translations.  Additionally, Table~\ref{tab:translation_instances - crosslingual} provides the number of sentence instances available for each translation process. The dataset is balanced across the different honorific levels, with the highest number of instances (1,412) observed in the Ngoko Alus$\leftrightarrow$ID pair. This distribution ensures that models are evaluated on a sufficiently diverse set of translation tasks.
\begin{table*}[h]
    \centering
    \large  
    \begin{adjustbox}{max width=\textwidth}
    \begin{tabular}{l|c|c|c|c}
    \toprule
    \multicolumn{1}{c|}{\textbf{Model}} & 
     \textbf{ID $\rightarrow$ Ngoko} & 
    \textbf{ID $\rightarrow$ Ngoko Alus} & 
     \textbf{ID $\rightarrow$ Krama} & 
    \textbf{ID $\rightarrow$ Krama Alus} \\
    \midrule
    GPT4o & \textbf{ 54.14} & \textbf{36.99} &  42.59 & \textbf{46.16} \\
    Gemini-1.5-pro &  47.49 & 32.33 & \textbf{ 45.55} & 31.66 \\
    Sailor2 8B &  26.50 & 20.36 &  17.99 & 19.85 \\
    Gemma2 9B Instruct &  11.85 & 11.45 &  10.98 & 11.03 \\
    Llama3.1 8B Instruct &  11.60 & 12.47 &  12.15 & 12.38 \\
    SahabatAI v1 Instruct (Llama3 8B) &  16.19 & 15.06 &  14.13 & 13.93 \\
    SahabatAI v1 Instruct (Gemma2 9B) &  31.02 & 25.68 &  21.47 & 23.26 \\
    \midrule
    \multicolumn{1}{c|}{\textbf{Model}} & 
     \textbf{Ngoko $\rightarrow$ ID} & 
    \textbf{Ngoko Alus $\rightarrow$ ID} & 
     \textbf{Krama $\rightarrow$ ID} & 
    \textbf{Krama Alus $\rightarrow$ ID} \\
    \midrule
    GPT4o & \textbf{ 61.23} & \textbf{53.25} & \textbf{ 56.61} & \textbf{51.49} \\
    Gemini-1.5-pro &  58.44 & 51.65 &  55.83 & 51.22 \\
    Sailor2 8B &  32.06 & 29.76 &  30.52 & 29.68 \\
    Gemma2 9B Instruct &  20.43 & 17.70 &  15.88 & 15.27 \\
    Llama3.1 8B Instruct &  12.99 & 12.39 &  12.46 & 12.37 \\
    SahabatAI v1 Instruct (Llama3 8B) &  29.07 & 26.66 &  25.97 & 25.10 \\
    SahabatAI v1 Instruct (Gemma2 9B) &  42.12 & 37.99 &  37.21 & 34.90 \\
    \bottomrule
    \end{tabular}
    \end{adjustbox}
    \caption{Cross-lingual evaluation using CHRF++ scores for translation between Indonesian (ID) and Javanese honorific levels.}
    \label{tab:performance_crosslingual_chrf}
\end{table*}

\begin{table}[h]
    \centering
    \resizebox{0.49\textwidth}{!}{
    \begin{tabular}{ll|c}
        \toprule
        \multicolumn{2}{c|}{Translation Pairs (Bidirectional)} & Number of Instances \\
        Label 1 & Label 2 & (Sentences) \\
        \midrule
        \midrule
        Ngoko & ID & 585 \\
        Ngoko Alus & ID & 1412 \\
        Krama & ID & 595 \\
        Krama Alus & ID & 584 \\
        \bottomrule
    \end{tabular}
    }
    \caption{Number of sentence instances for each translation process. Each translation process is bidirectional, meaning the count for a given pair (e.g., ID $\leftrightarrow$ Ngoko) is identical in both directions.}
    \label{tab:translation_instances - crosslingual}
\end{table}

\section{Computational Resources \& Hyper-parameters}
We utilized four A40 40GB GPUs for the experiments related to downstream tasks.
\label{sec:hyper-parameters}

\subsection{Honorific Level Classification}
The model is configured with temperature 0.1, top-p 0.9, and a maximum of 50 new tokens. A low temperature (0.1) ensures highly deterministic outputs, which is crucial for accurate classification. The top-p (0.9) parameter helps retain relevant token probabilities while allowing slight flexibility in word selection. The max\_new\_tokens (50) limit prevents unnecessary token generation, ensuring concise and precise classification.

\subsection{Honorific Style Changes}
For translating from one honorific level to another, the model generates a single output sequence to maintain consistency. A temperature of 0.7 ensures a balance between randomness and coherence, while top-p (0.9) and top-k (50) filtering help retain relevant word choices. The repetition penalty (1.2) discourages redundant text generation, and early stopping prevents unnecessary token extension. The min\_length (10) and max\_new\_tokens (100) parameters control the length of generated text, ensuring meaningful responses without excessive verbosity.

\subsection{Honorific Cross-lingual Translation}
When translating between Javanese and Indonesian across different honorific levels, the hyperparameters remain consistent to preserve translation fairness. The model generates a single sequence while using temperature (0.7) and top-p (0.9) to maintain lexical variety. The top-k (50) value helps focus on relevant words while avoiding excessive randomness. Early stopping enhances efficiency, and a repetition penalty (1.2) ensures fluency. The minimum length constraint (10) prevents overly short outputs, while the maximum token limit (100) maintains reasonable translation length.

\subsection{Conversation Generation}
For generating honorific-level conversations in Javanese, the hyperparameters are set to balance fluency and creativity. The model generates one response per prompt, while temperature (0.7) introduces controlled randomness to simulate natural conversations. Top-p (0.9) and top-k (50) work together to refine token selection, preventing the model from choosing unlikely continuations. The repetition penalty (1.2) avoids repetitive responses, and early stopping enhances efficiency. The length constraints (min\_length = 10, max\_new\_tokens = 100) help maintain coherent and meaningful dialogue structures.

\subsection{LSTM}
The model's architecture includes an embedding layer with a vocabulary size of 10,000 and an embedding size of 64. Input sequences are limited to a maximum length of 128. A convolutional layer with 32 filters and a kernel size of 3 is applied, followed by a max pooling layer to reduce dimensionality. The key component is a bidirectional LSTM with 32 units, capturing sequential dependencies from both directions. This architecture, which combines a bidirectional LSTM with convolutional and max-pooling layers, has proven effective in text classification tasks \citep{LIU2019325}. A dropout layer with a rate of 0.4 is used to prevent overfitting. The final output layer is a dense layer with 4 units, representing the honorific levels, and uses softmax activation. The model is trained using the Adam optimizer and categorical cross-entropy loss, with early stopping based on validation accuracy to avoid overfitting. Training is conducted for up to 100 epochs with a batch size of 32.

\subsection{Finetuned Models}
The fine-tuned models demonstrated robust performance across all evaluation metrics. Table \ref{tab:hyper} details the hyper-parameters used in these experiments, revealing that a learning rate of 5e-5 and a batch size of 16 consistently produced strong results across models. The hyper-parameter tuning was conducted on a validation subset to refine the candidate of classifier models' performance. All fine-tuning experiments were conducted with a fixed random seed (set to 42) to ensure consistent results across runs.



\section{Others}

\subsection{Shannon Entropy}
\label{app:shannon}
\begin{equation}
\label{eq:shannon_entropy}
H = - \sum_{i=1}^{n} p_i\log_2(p_i),
\end{equation}
where $p_i$ is probability of a specific honorific form.

\subsection{Yule's characteristic K value}
\begin{equation}
\small
K = 10^4 \times \frac{\sum\limits_{m} \left[m^2 V(m, N)\right] - N}{N^2}
\label{eq:yule-k}
\end{equation}
where $N$ represents the total number of word tokens, and $V(m, N)$ is a function that calculates the number of word types occurring exactly $m$ times in the corpus.

\subsection{Kullback-Leibler Divergence}
\label{app:kl}
\begin{equation}
\label{eq:kl-divergence}
\scalebox{0.68}{$
    \text{KL}_{\text{sym}}(P, Q) = \frac{1}{2} \sum_i P(i) \log \frac{P(i)}{Q(i)} + \frac{1}{2} \sum_i Q(i) \log \frac{Q(i)}{P(i)}
$}
\end{equation}
where $P(i)$ and $Q(i)$ denote the probability of token $i$ in the Javanese and Indonesian distributions, respectively. We apply Laplace smoothing to avoid zero probabilities and use a symmetric version to ensure comparability across directions.

\subsection{Jensen Score}
\label{app:jensen}
\begin{equation}
\label{eq:jensen-score}
\scalebox{0.58}{$
    \text{JSD}(P \parallel Q) = \frac{1}{2} \text{KL}(P \parallel M) + \frac{1}{2} \text{KL}(Q \parallel M), \quad \text{where } M = \frac{1}{2}(P + Q)
$}
\end{equation}
where $P$ and $Q$ are the token distributions of Javanese and Indonesian sentences, and $M$ is the average distribution. The Jensen score corresponds to the Jensen-Shannon divergence and is computed as the squared distance output from \texttt{scipy}'s \texttt{jensenshannon} function.

\subsection{Rule Based Algorithm}
\label{app:rulebasedalgo}
Algorithm~\ref{alg:javanese_classification} classifies a given Javanese sentence into one of four honorific levels: \textit{Ngoko} (informal), \textit{Ngoko Alus} (polite informal), \textit{Krama} (formal), or \textit{Krama Alus} (highly formal). It first counts the occurrences of words associated with each speech level and calculates their proportions relative to the total words in the sentence. The classification is determined based on the dominant proportion, with ties defaulting to \textit{Ngoko}. Special words from the \textit{Krama Inggil} lexicon can upgrade the classification, while the presence of \textit{Ngoko} words can downgrade it, ensuring a more nuanced classification.

\subsection{Javanese Language Dataset}
\label{app:datasetstats}
Table \ref{tab:javanese_dataset} presents an overview of various Javanese language corpora, detailing their distribution across four honorific levels: \textit{Ngoko}, \textit{Ngoko Alus}, \textit{Krama}, and \textit{Krama Alus}. The dataset sizes vary significantly, ranging from 105 to over 185,000 entries, reflecting diverse sources and linguistic compositions. Additionally, Shannon’s Entropy is computed for each corpus to quantify variability in honorific usage, where higher entropy values indicate a more balanced distribution across honorific levels, suggesting a richer representation of Javanese linguistic diversity. Notably, the dataset \textit{\datasetname} exhibits the highest entropy (1.8763), implying a well-balanced honorific distribution, while corpora such as \textit{The Identifikasi-Bahasa} and \textit{OSCAR-2301 Javanese} display lower entropy, indicating a more skewed distribution. The data presented in this table corresponds to the visualization depicted in Figure \ref{fig:dataset_coverage}, which illustrates the honorific level distributions and entropy values across different corpora.

\begin{table*}[h]
\centering
\begin{tabular}{lcccccc}
\toprule
\textbf{Model} & \textbf{lr} & \textbf{Batch Size} & \textbf{Acc.} & \textbf{Prec.} & \textbf{Rec.} & \textbf{F1} \\
\midrule
Javanese BERT & 5e-5 & 16 & 93.01 & 92.98 & 93.01 & 92.97 \\
 & 5e-5 & 32 & \textbf{93.19} & \textbf{93.16} & \textbf{93.19} & \textbf{93.16} \\
 & 2e-5 & 16 & 92.11 & 92.15 & 92.11 & 92.12 \\
 & 2e-5 & 32 & 91.03 & 91.09 & 91.03 & 91.06 \\
 & 1e-5 & 16 & 89.78 & 89.70 & 89.78 & 89.73 \\
 & 1e-5 & 32 & 81.36 & 80.49 & 81.36 & 78.97 \\
Javanese GPT & 5e-5 & 16 & \textbf{92.65} & \textbf{92.67} & \textbf{92.65} & \textbf{92.65} \\
 & 5e-5 & 32 & 89.96 & 90.18 & 89.96 & 89.96 \\
 & 2e-5 & 16 & 86.38 & 86.56 & 86.38 & 86.37 \\
 & 2e-5 & 32 & 82.62 & 83.28 & 82.62 & 82.83 \\
 & 1e-5 & 16 & 80.47 & 80.13 & 80.47 & 80.18 \\
 & 1e-5 & 32 & 74.55 & 73.92 & 74.55 & 73.34 \\
Javanese DistilBERT & 5e-5 & 16 & \textbf{93.19} & \textbf{93.12} & \textbf{93.19} & \textbf{93.13} \\
 & 5e-5 & 32 & 91.40 & 91.29 & 91.40 & 91.32 \\
 & 2e-5 & 16 & 88.35 & 88.09 & 88.35 & 88.15 \\
 & 2e-5 & 32 & 86.20 & 86.39 & 86.20 & 85.28 \\
 & 1e-5 & 16 & 79.57 & 79.33 & 79.57 & 74.65 \\
 & 1e-5 & 32 & 73.30 & 65.61 & 73.30 & 63.64 \\
\bottomrule
\end{tabular}
\caption{\label{tab:hyper}
Performance comparison of different models with hyper-parameter tuning conducted on the validation subset of the Javanese Honorific Corpus. While the \textbf{javanese-bert-small-imdb-classifier} model achieved the best overall results with a batch size of 32, the performance with a batch size of 16 was only slightly lower. Across all models, a learning rate of 5e-5 and a batch size of 16 consistently provided strong results, making this the most stable hyper-parameter combination.}
\end{table*}

\begin{algorithm*}[h]
\algsetup{linenosize=\small}
\small  
\caption{Rule-Based Classification of Javanese Speech Levels}
\label{alg:javanese_classification}
\begin{algorithmic}[1]
\REQUIRE A sentence $s$ to classify
\ENSURE Classification label $l \in \{0,1,2,3\}$ where:
        \\[0.2em] \hspace{1em} 0: Ngoko
        \\[0.2em] \hspace{1em} 1: Ngoko Alus
        \\[0.2em] \hspace{1em} 2: Krama
        \\[0.2em] \hspace{1em} 3: Krama Alus

\medskip
\STATE \textbf{Initialize word counts:}
\STATE $w_{\text{ngoko}} \gets \text{count\_words}(s, \text{``ngoko''})$
\STATE $w_{\text{krama}} \gets \text{count\_words}(s, \text{``krama''})$
\STATE $w_{\text{kramaAlus}} \gets \text{count\_words}(s, \text{``krama\_alus''})$
\STATE $w_{\text{kramaInggil}} \gets \text{count\_words}(s, \text{``krama\_inggil''})$
\STATE $w_{\text{total}} \gets w_{\text{ngoko}} + w_{\text{krama}} + w_{\text{kramaAlus}} + w_{\text{kramaInggil}}$

\medskip
\STATE \textbf{Calculate proportions:}
\STATE $p_{\text{ngoko}} \gets w_{\text{ngoko}}/w_{\text{total}}$
\STATE $p_{\text{krama}} \gets w_{\text{krama}}/w_{\text{total}}$
\STATE $p_{\text{kramaAlus}} \gets w_{\text{kramaAlus}}/w_{\text{total}}$

\medskip
\STATE \textbf{Initial classification:}
\IF{$p_{\text{ngoko}} > \max(p_{\text{krama}}, p_{\text{kramaAlus}})$}
    \STATE $l \gets 0$ \COMMENT{Ngoko}
\ELSIF{$p_{\text{krama}} > \max(p_{\text{ngoko}}, p_{\text{kramaAlus}})$}
    \STATE $l \gets 2$ \COMMENT{Krama}
\ELSE
    \STATE $l \gets 3$ \COMMENT{Krama Alus}
\ENDIF

\medskip
\STATE \textbf{Handle ties:}
\IF{$p_{\text{ngoko}} = p_{\text{kramaAlus}}$}
    \STATE $l \gets 0$ \COMMENT{Default to Ngoko}
\ENDIF

\medskip
\STATE \textbf{Adjust based on special words:}
\IF{$w_{\text{kramaInggil}} > 0$}
    \STATE $l \gets \min(l + 1, 3)$ \COMMENT{Upgrade level}
\ENDIF
\IF{$w_{\text{ngoko}} > 0$}
    \STATE $l \gets \max(l - 1, 0)$ \COMMENT{Downgrade level}
\ENDIF

\medskip
\RETURN{$l$}
\end{algorithmic}
\end{algorithm*}

\begin{table*}[h]
    \centering
    \resizebox{0.98\textwidth}{!}{
    \begin{tabular}{l|cccc|c|c}
        \toprule
         & \multicolumn{4}{c|}{\textbf{Honorific Level's Distribution Percentage}} & & \textbf{Shannon} \\
        \textbf{Dataset} & \textbf{Ngoko (\%)} & \textbf{Ngoko Alus (\%)} & \textbf{Krama (\%)} & \textbf{Krama Alus (\%)} & \textbf{Size} & \textbf{Entropy} \\
        \midrule
        \textbf{$\datasetname$} & 35.26 & 14.66 & 35.14 & 14.94 & 4024 & 1.87 \\
        The identifikasi-bahasa & 96.05 & 0.93 & 2.91 & 0.11 & 44109 & 0.27 \\
        Javanese dialect identification & 92.09 & 2.21 & 4.53 & 1.17 & 16498 & 0.50 \\
        Korpus-Nusantara (Jawa) & 9.00 & 1.00 & 86.40 & 3.60 & 1000 & 0.73 \\
        Korpus-Nusantara (Jawa Ngoko) & 82.59 & 9.95 & 4.49 & 2.97 & 6059 & 0.91 \\
        JV-ID-ASR & 74.80 & 3.90 & 19.49 & 1.81 & 185076 & 1.06 \\
        JV-ID TTS (female) & 66.97 & 1.36 & 27.41 & 4.26 & 2864 & 1.17 \\
        JV-ID TTS (male) & 67.58 & 1.15 & 27.08 & 4.19 & 2958 & 1.15 \\
        OSCAR-2301 Javanese & 95.24 & 0.95 & 3.81 & 0.00 & 105 & 0.31 \\
        Jvwiki & 89.20 & 2.95 & 6.23 & 1.62 & 37335 & 0.64 \\
        \bottomrule
    \end{tabular}
    }
    \caption{Javanese Language Corpus Statistics}
    \label{tab:javanese_dataset}
\end{table*}

\vspace{-0.5em}
\input{prompts/task1}

\vspace{-0.5em}
\input{prompts/task2}

\vspace{-0.5em}
\input{prompts/task3}

\vspace{-0.5em}
\input{prompts/task4noHint}

\vspace{-0.5em}
\input{prompts/task4hint}

\clearpage

\raggedright

\end{document}

%% file: prompts/task1.tex
\begin{figure*}[h]
\centering
\begin{tcolorbox}[enhanced, colback=blue!5!white, colframe=blue!75!black,
  title=Prompt example for Task 1: Honorific Level Classification, width=\textwidth]
Analyze the Javanese sentence enclosed in square brackets.\\
Determine if it is ngoko, ngoko alus, krama, or krama alus.\\
Return the answer as the corresponding text label: 0 (ngoko), 1 (ngoko alus), 2 (krama), 3 (krama alus).\\
Provide only the integer label without any additional explanation. [\textcolor{red}{<SENTENCE>}]
\end{tcolorbox}
\caption{Honorific level classification task's prompt.}
\label{fig:prompt-task1-example}
\end{figure*}

%% file: prompts/task2.tex
\begin{figure*}[h]
\centering
\begin{tcolorbox}[enhanced, colback=blue!5!white, colframe=blue!75!black,
  title=Prompt example for Task 2: Honorific Style Change, width=\textwidth]
Translate this Javanese sentence from \textcolor{red}{<HONORIFIC LEVEL A>} to \textcolor{red}{<HONORIFIC LEVEL B>}:\\
Source: \textcolor{red}{<TEXT A>}\\
Target:
\end{tcolorbox}
\caption{Honorific style change task’s prompt.}
\label{fig:prompt-task2-example}
\end{figure*}

%% file: prompts/task3.tex
\begin{figure*}[h]
\centering
\begin{tcolorbox}[enhanced, colback=blue!5!white, colframe=blue!75!black,
  title=Prompt example for Task 3: Cross-lingual Honorific Translation, width=\textwidth]
Translate this Indonesian sentence to Javanese using \textcolor{red}{<HONORIFIC LEVEL>}:\\
Indonesian: \textcolor{red}{<INDONESIAN TEXT>}\\
Javanese:
\tcblower
Translate this Javanese sentence to Indonesian:\\
Javanese: \textcolor{red}{<JAVANESE TEXT>}\\
Indonesian:
\end{tcolorbox}
\caption{Cross-lingual honorific translation task’s prompt.}
\label{fig:prompt-task3-example}
\end{figure*}

%% file: prompts/task4noHint.tex
\begin{figure*}[h]
\centering
\begin{tcolorbox}[enhanced, colback=blue!5!white, colframe=blue!75!black,
  title=Prompt example for Task 4: Conversation Generation (without hint), width=\textwidth]
Create a conversation between A as \textcolor{red}{<ROLE A>} and B as \textcolor{red}{<ROLE B>} in Javanese language with this context: `\textcolor{red}{<CONTEXT>}` \\
Please follow this format: \\
A: `<UTTERANCE>` \\
B: `<UTTERANCE>` \\ \\
Answer:
\end{tcolorbox}
\caption{Coversation generation task's prompt (without hint).}
\label{fig:prompt-task4-nohint-example}
\end{figure*}

%% file: prompts/task4hint.tex
\begin{figure*}[h]
\centering
\begin{tcolorbox}[enhanced, colback=blue!5!white, colframe=blue!75!black,
  title=Prompt example for Task 4: Conversation Generation (with hint), width=\textwidth]
Create a conversation between A as \textcolor{red}{<ROLE A>} and B as \textcolor{red}{<ROLE B>} in Javanese language with this context: `\textcolor{red}{<CONTEXT>}` \\
Please follow this format: \\
A: `<UTTERANCE>` \\
B: `<UTTERANCE>` \\ \\
Use this Javanese's honorific level usage as a hint:\\
1. Ngoko:\\
- Used for informal conversations with peers or lower-status individuals\\
- Common in close relationships or familiar interactions\\
2. Ngoko Alus:\\
- Adds respect when speaking to equals or higher-status individuals in informal or close relationships\\
- Flexible for conversations with mixed-status participants\\
- Talked with a person with equals status about other person who has a higher-status
3. Krama:\\
- Used for respectful conversations with equals or higher-status individuals, especially when not close\\
- Suitable for maintaining formality in less familiar interactions\\
4. Krama Alus:\\
- Expresses high respect in conversations with higher-status individuals or unfamiliar equals\\
- Essential for formal interactions requiring utmost politeness\\
- Talked with a person with higher-status about other person who has a higher-status\\ \\
Answer:
\end{tcolorbox}
\caption{Coversation generation task's prompt (with hint).}
\label{fig:prompt-task4-hint-example}
\end{figure*}